\documentclass[10pt,twocolumn,letterpaper]{article}

\usepackage{cvpr}

\usepackage{CJK}
\usepackage{listings}

\usepackage{hyperref}       
\usepackage{url}            
\usepackage{booktabs}       
\usepackage{amsfonts}       
\usepackage{nicefrac}       
\usepackage{microtype}      
\usepackage{xcolor}         
\usepackage{amsmath}
\usepackage{booktabs}
\usepackage{url}            
\usepackage{booktabs}       
\usepackage{xspace}
\usepackage{nicefrac}       
\usepackage{microtype}      
\usepackage{xcolor}         
\usepackage{graphicx}
\usepackage{url}            
\usepackage{booktabs}       
\usepackage{amsfonts}       
\usepackage{nicefrac}       
\usepackage{microtype}      
\usepackage{xcolor}         
\usepackage{wasysym}
\usepackage{multirow}
\usepackage{tcolorbox}
\usepackage{amsfonts}
\usepackage{pifont}
\usepackage{fontawesome}
\usepackage{float}
 \usepackage{enumitem}
\usepackage[dvipsnames]{xcolor}
\usepackage[normalem]{ulem}
\usepackage{xspace}
\usepackage{makecell}
\usepackage{wrapfig}
\usepackage{xcolor}
\usepackage{colortbl}
\usepackage{multirow}
\usepackage[table]{xcolor}
\usepackage{tabularx}
\usepackage{tcolorbox}

\definecolor{cvprblue}{rgb}{0.21,0.49,0.74}
\hypersetup{
    colorlinks=true,
    allcolors=cvprblue
}
\definecolor{softblue}{RGB}{230,245,255}
\definecolor{softgreen}{RGB}{220,245,220}
\definecolor{softyellow}{HTML}{FFF2CC}
\definecolor{softpurple}{RGB}{245,235,255}
\definecolor{softorange}{RGB}{255,235,210}
\definecolor{softpink}{RGB}{255,230,240}
\definecolor{lightgray}{RGB}{240,240,240}

\def\paperID{*****}
\def\confName{CVPR}
\def\confYear{2026}

\newcommand{\Description}[1]{}

\title{LongAV-Compass: Towards Unified Evaluation of Minute-Scale Audio-Visual Generation Across T2AV, I2AV, and V2AV}

\author{
    Tengfei Liu$^{1}$\,
    Yang Shi$^{1,2}$\thanks{Corresponding Author} \,\,
    Xuanyu Zhu$^{1}$\,
    Jiafu Tang$^{3}$\,
    Liu Yang$^{4}$\,
    Qixun Wang$^{1}$\,
    Zhuoran Zhang$^{1}$\,
    \\
    Yuqi Tang$^{5}$\,
    Fengxiang Wang$^{6}$\,
    Yuhao Dong$^{7}$\,
    Xinlong Chen$^{8}$\,
    Bozhou Li$^{1}$\,
    Bohan Zeng$^{1}$\,
    Yue Ding$^{8}$\,
    \\
    Xiaohan Zhang$^{3}$\,
    Jialu Chen$^{2}$\,
    Haotian Wang$^{9}$\footnotemark[1] \,
    Yuanxing Zhang$^{2}$\thanks{Project Lead} \,
    Pengfei Wan$^{2}$\,
    Leye Wang$^{1}$\footnotemark[1]\,
    \\ 
    $^1$Peking University\quad
    $^2$Kling Team\quad
    $^3$Nanjing University\quad
    $^4$SJTU\quad
    $^5$HKUST(GZ)\quad
    \\
    $^6$Shanghai AI Lab\quad
    $^7$Nanyang Technological University\quad
    $^8$CASIA\quad
    $^9$Tsinghua University\quad
    \\
    {\centering}
    \url{https://github.com/pkucs-Ltf/LongAV-Compass}
}

\newcommand{\name}{LongAV-Compass\xspace}

\begin{document}
\maketitle

\begin{abstract}
Audio-visual generation is rapidly advancing from short clips to minute-long content, while existing evaluation protocols remain largely confined to short-form settings. 
Existing benchmarks primarily focus on 5--10 second text-conditioned generation and rarely support unified evaluation across text, image, and video conditioning modalities. 
Moreover, they provide limited insight into how identity consistency, narrative coherence, and audio-visual alignment degrade over extended temporal horizons.
To bridge this gap, we introduce \textbf{\name}, a systematic benchmark for minute-long audio-visual generation. \name contains 284 curated test cases spanning text-to-audio-video (T2AV), image-to-audio-video (I2AV), and video-to-audio-video (V2AV), organized by application scenario and generation complexity. 
The benchmark combines taxonomy-guided benchmark construction with a unified evaluation framework that integrates MLLM-assisted assessment with complementary perceptual and multimodal metrics, including DINO-v2, ArcFace, CLIP, and ImageBind.
The framework evaluates more than 20 fine-grained dimensions covering within-segment quality, cross-segment consistency, global narrative coherence, semantic alignment, and audio-visual synchronization.
Through experiments on 11 representative models together with human-alignment validation, \name provides a diagnostic testbed for analyzing the limitations of current systems in sustaining coherent, semantically aligned, and temporally consistent minute-scale audio-visual generation across diverse input modalities.
\end{abstract}

\noindent\textbf{Keywords:} Audio-Visual Generation, Long Video Generation, Evaluation

\section{Introduction}
\label{sec:introduction}

\begin{figure*}[!t]
  \centering
  \includegraphics[width=\textwidth]{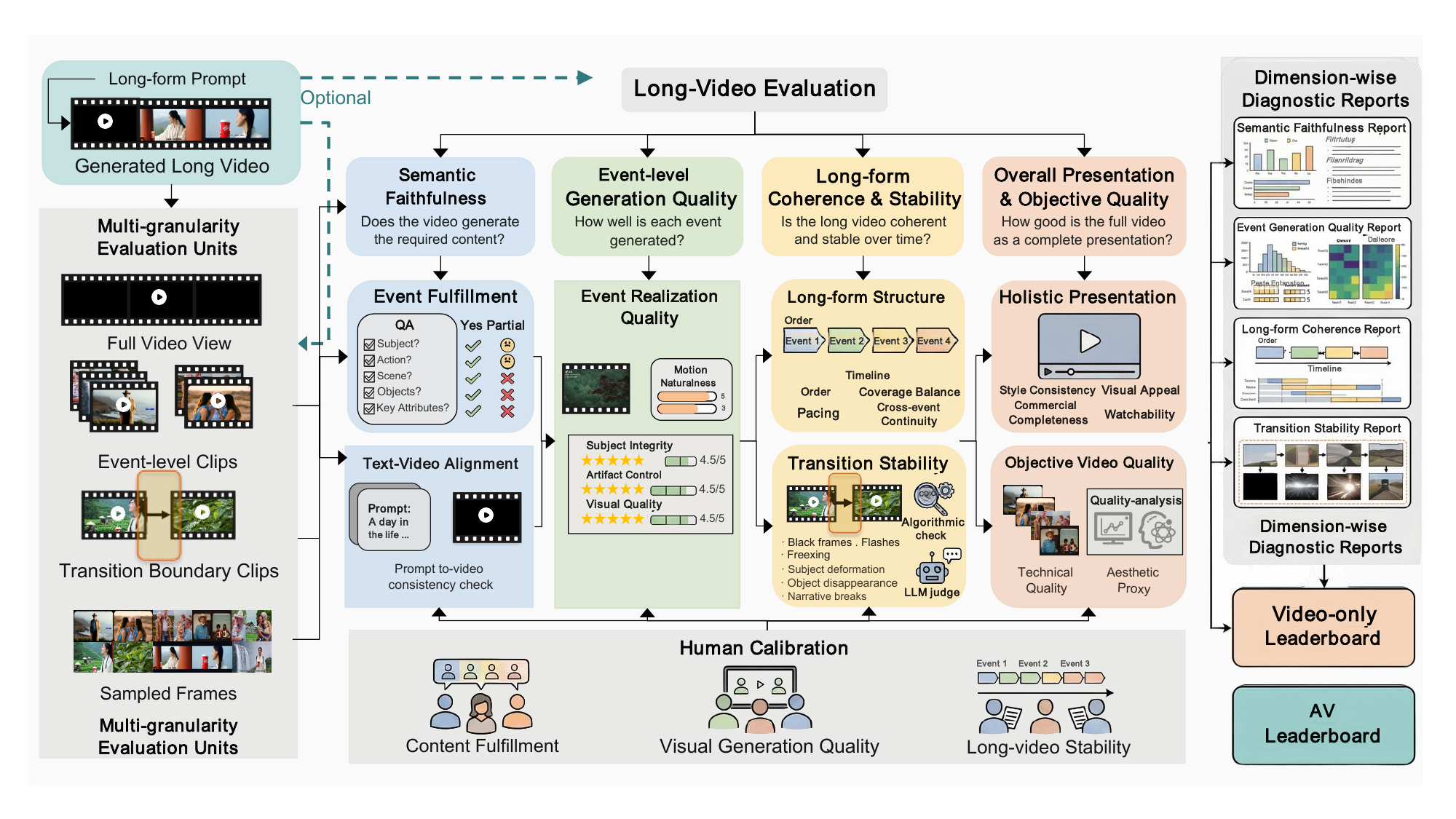}
  \caption{\textbf{Overview of \name}. The benchmark unifies T2AV, I2AV, and V2AV under shared taxonomy, event-level annotation, and a hierarchical evaluation framework, enabling diagnosis of long-range audio-visual failures beyond flat leaderboard comparison.}
  \label{fig:overview}
  \Description{A four-part benchmark overview diagram showing task inputs, benchmark design, unified evaluation framework, and outputs with diagnosis modules for \name.}
\end{figure*}

Recent advances in video generation models are pushing audio-visual generation beyond short clips. 
Commercial and open-source systems increasingly support longer durations, richer prompting, and native or compositional audio generation, making minute-scale outputs relevant to applications such as vlogs, tutorials, product demonstrations, advertisements, and story-driven content. 
In this setting, success is no longer determined by producing a visually plausible 5-second clip. 
Instead, models must sustain subject identity, event continuity, scene transitions, and audio grounding over substantially longer temporal horizons.

However, evaluation has not kept pace with this shift. 
Existing benchmarks for video and audio-visual generation remain largely focused on short-form settings, where a single clip is often sufficient to assess local visual quality or coarse semantic alignment. 
Benchmarks such as VBench~\cite{huang2023vbench} and EvalCrafter~\cite{liu2023evalcrafter} have advanced standardized evaluation for video generation models, while recent audio-visual benchmarks such as VABench~\cite{hua2025vabench} and T2AV-Compass~\cite{cao2025t2avcompass} further extend evaluation to synchronized audio generation. 
These benchmarks provide valuable tools for short-video assessment, but their design does not fully capture the challenges of long-form generation, where failures often emerge only across multiple events, larger temporal gaps, or prolonged audio-visual interactions.

This gap leads to three key limitations. 
First, current benchmarks operate at a temporal scale that provides limited evidence about whether models can remain coherent over minute-long generation. 
Second, their coverage is often fragmented across input conditions, making it difficult to compare text-to-audio-video (T2AV), image-to-audio-video (I2AV), and video-to-audio-video (V2AV) systems under a unified protocol. 
Third, current evaluation offers limited diagnostic visibility into long-range degradation, such as cross-event identity drift, weak continuation quality, unstable scene transitions, and the decay of audio-visual synchronization as duration increases.

\begin{table*}[t]
\centering
\caption{\textbf{Comparisons between \name and representative video and audio-visual generation benchmarks.} \name focuses on two missing axes in prior evaluation: unified X2AV coverage across T2AV, I2AV, and V2AV, and longer sample duration. Here, V2A denotes generating audio for a given video, whereas V2AV evaluates video-conditioned audio-video continuation. The final column indicates whether the benchmark explicitly reports an average sample or video duration exceeding one minute.}
\label{tab:benchmark_comparison}
\scriptsize
\resizebox{\textwidth}{!}{
\begin{tabular}{lcccccccc}
\toprule
\textbf{Benchmark} & \textbf{\#Samples} & \textbf{T2V} & \textbf{T2AV} & \textbf{I2AV} & \textbf{V2A} & \textbf{V2AV} & \textbf{Unified X2AV} & \textbf{Avg. Video Duration $>$ 1min} \\
\midrule
MSVBench~\cite{shi2026msvbench} & 276 & \textcolor{ForestGreen}{\ding{51}} & \textcolor{red}{\ding{55}} & \textcolor{red}{\ding{55}} & \textcolor{red}{\ding{55}} & \textcolor{red}{\ding{55}} & \textcolor{red}{\ding{55}} & \textcolor{red}{\ding{55}} \\
AVGen-Bench~\cite{zhou2026avgenbench} & 235  & \textcolor{red}{\ding{55}} & \textcolor{ForestGreen}{\ding{51}} & \textcolor{red}{\ding{55}} & \textcolor{red}{\ding{55}} & \textcolor{red}{\ding{55}} & \textcolor{red}{\ding{55}} & \textcolor{red}{\ding{55}} \\
T2AV-Compass~\cite{cao2025t2avcompass} & 500  & \textcolor{red}{\ding{55}} & \textcolor{ForestGreen}{\ding{51}} & \textcolor{red}{\ding{55}} & \textcolor{red}{\ding{55}} & \textcolor{red}{\ding{55}} & \textcolor{red}{\ding{55}} & \textcolor{red}{\ding{55}} \\
VABench~\cite{hua2025vabench} & 1,299  & \textcolor{red}{\ding{55}} & \textcolor{ForestGreen}{\ding{51}} & \textcolor{ForestGreen}{\ding{51}} & \textcolor{red}{\ding{55}} & \textcolor{red}{\ding{55}} & \textcolor{red}{\ding{55}} & \textcolor{red}{\ding{55}} \\
PhyAVBench~\cite{xie2025phyavbench} & 337  & \textcolor{red}{\ding{55}} & \textcolor{ForestGreen}{\ding{51}} & \textcolor{ForestGreen}{\ding{51}} & \textcolor{ForestGreen}{\ding{51}} & \textcolor{red}{\ding{55}} & \textcolor{red}{\ding{55}} & \textcolor{red}{\ding{55}} \\
VinTAGe-Bench~\cite{kushwaha2025vintage} & 636 & \textcolor{red}{\ding{55}} & \textcolor{red}{\ding{55}} & \textcolor{red}{\ding{55}} & \textcolor{ForestGreen}{\ding{51}} & \textcolor{red}{\ding{55}} & \textcolor{red}{\ding{55}} & \textcolor{red}{\ding{55}} \\
\midrule
\textbf{\name} & \textbf{284} & \textcolor{red}{\ding{55}} & \textcolor{ForestGreen}{\ding{51}} & \textcolor{ForestGreen}{\ding{51}} & \textcolor{red}{\ding{55}} & \textcolor{ForestGreen}{\ding{51}} & \textcolor{ForestGreen}{\ding{51}} & \textcolor{ForestGreen}{\ding{51}} \\
\bottomrule
\end{tabular}}
\vspace{1mm}
\footnotesize
\end{table*}

As summarized in Table~\ref{tab:benchmark_comparison}, existing benchmarks typically cover only part
of the X2AV task space or remain focused on short-form generation, leaving unified minute-scale audio-visual evaluation underexplored.

To address these limitations, we introduce \textbf{\name}, a unified benchmark for minute-scale audio-visual generation. 
\name contains $284$ curated test cases, including $128$ T2AV examples, $115$ I2AV examples, and $41$ V2AV examples. 
The benchmark is organized according to a two-dimensional taxonomy of application scenario and generation complexity, covering Vlog, Content-Creator, Performance Ads, and Brand Ads. 
Each test case is annotated with both a global description and event-level structure, enabling evaluation of long-form narrative organization rather than isolated frames or short clips.

Beyond dataset construction, \textbf{\name} provides a unified evaluation framework tailored to long-form audio-visual generation. 
The framework assesses more than $20$ fine-grained dimensions spanning within-segment video quality, cross-segment consistency, global narrative coherence, long-audio quality, audio-visual synchronization, and input-conditioned semantic alignment. 
It follows an MLLM-centered evaluation protocol based on Gemini 3.1 Pro~\cite{gemini31pro}, complemented by specialized perceptual and multimodal metrics including DINO-v2~\cite{oquab2023dinov2} and CLIP~\cite{radford2021learning}. 
This hybrid design enables evaluation from complementary perspectives, including segment-level quality, cross-segment subject consistency, script following, semantic alignment, image anchoring, video continuation quality, and audio-visual synchronization. 
We further conduct a human-alignment study to validate the reliability of the resulting scores.

Figure~\ref{fig:overview} illustrates the overall design of \textbf{\name}. 
It unifies T2AV, I2AV, and V2AV under a shared taxonomy, event-level annotation schema, and hierarchical evaluation framework, while still supporting task-specific diagnostics and leaderboards. 
Rather than serving as a simple extension of short-form leaderboards, \name is designed as a diagnostic benchmark for understanding long-form audio-visual generation. 
Through unified evaluation of $11$ representative systems, it enables systematic analysis of model capabilities and failure modes, including long-range identity drift, brittle event transitions, conditioning-specific weaknesses, and unstable minute-scale audio continuity.

Our contributions are summarized as follows:
\begin{itemize}
\item We introduce \textbf{\name}, the first benchmark dedicated to minute-scale audio-visual generation across text, image, and video inputs, with $284$ curated test cases organized by application scenario and generation complexity.
\item We design a unified evaluation framework for long-form audio-visual generation across T2AV, I2AV, and V2AV. The framework evaluates more than $20$ dimensions and decomposes long-video assessment into three complementary perspectives: within-segment quality, cross-segment consistency, and global narrative coherence, together with audio-visual synchronization and input-conditioned semantic alignment.
\item We conduct a comprehensive evaluation of $11$ representative generation systems under the proposed protocol. Beyond overall ranking, our analysis reveals the capabilities current models handle well and the failure modes they still exhibit, providing a systematic diagnosis of long-form audio-visual generation.
\end{itemize}

\section{Related Work}
\label{sec:related_work}

\subsection{Benchmarks on Short-Form Video Generation}
Progress in benchmarking video generation has been largely driven by short-form evaluation suites such as VBench~\cite{huang2023vbench}, EvalCrafter~\cite{liu2023evalcrafter}, and FETV~\cite{liu2023fetv}. 
These benchmarks define systematic evaluation dimensions covering visual quality, motion realism, semantic alignment, and prompt following~\cite{huang2025vbench++,feng2024tc,sun2025ve}, enabling more standardized comparisons among video generation models. 
However, their protocols are primarily designed for short text-conditioned clips, making them less suitable for assessing long-form audio-visual generation. 
In particular, they provide limited evidence about whether models can preserve subject identity, narrative coherence, scene continuity, and audio-visual consistency over minute-long outputs, where failures may accumulate across multiple events rather than appear within a single short clip.

\subsection{Benchmarks on Audio-Visual Generation}

Recent studies have extended generative evaluation from video-only generation to synchronized audio-video synthesis. 
In parallel, audio-video generation models have explored joint multimodal
generation, as in MM-Diffusion~\cite{ruan2023mmdiffusion},
VideoPoet~\cite{kondratyuk2023videopoet}, and Movie
Gen~\cite{meta2024moviegen}, while video-to-audio methods such as Diff-
Foley~\cite{luo2023difffoley}, FoleyCrafter~\cite{zhang2024foleycrafter}, and
STA-V2A~\cite{ren2024stav2a} focus on temporally and semantically aligned
sound generation for videos.
VABench~\cite{hua2025vabench} introduces a multi-dimensional benchmark for audio-video generation across multiple task types, while T2AV-Compass~\cite{cao2025t2avcompass} proposes a unified evaluation protocol for text-to-audio-video systems. 
These efforts broaden evaluation beyond visual quality and reveal important limitations of current audio-video generation models. 
Nevertheless, they remain primarily focused on short-form generation and do not systematically examine long-range challenges in minute-scale content, such as cross-event consistency degradation, audio-visual synchronization decay, and input-conditioned continuation across text, image, and video modalities.

\subsection{Story-Level and Long-Horizon Evaluation}
StoryBench~\cite{bugliarello2023storybench} extends evaluation beyond single-sentence prompting by introducing temporally structured assessment for continuous story visualization, while recent multi-shot benchmarks such as MSVBench~\cite{shi2026msvbench} further emphasize hierarchical scripts and cross-shot consistency.
By emphasizing event sequences and story coherence, StoryBench represents an important step toward long-horizon generative evaluation. 
However, it focuses on text-conditioned story visualization rather than minute-long audio-visual generation, and does not address reference-image conditioning, reference-video continuation, or long-range audio assessment.

Overall, prior benchmarks have advanced short-form video evaluation, audio-visual generation assessment, and story-level generation analysis from complementary perspectives. 
In contrast, \textbf{\name} targets a distinct evaluation regime: minute-long audio-visual generation across T2AV, I2AV, and V2AV, with taxonomy-guided coverage and a unified evaluation framework designed to diagnose long-range consistency, event-level continuity, and cross-modal alignment as duration and structure increase.

\section{\name}
\label{sec:benchmark}

\subsection{Task Formulation}

\begin{table}[t]
    \caption{\textbf{Task coverage in \name.} \texttt{S}, \texttt{RI}, and \texttt{RV} denote script, reference image, and reference video, respectively.}
    \label{tab:task_coverage}
    \centering
    \begin{tabularx}{\linewidth}{
    p{0.10\linewidth}
    p{0.18\linewidth}
    p{0.18\linewidth}
    p{0.18\linewidth}
    X}
    \toprule
    \textbf{Task} & \textbf{\#Samples} & \textbf{\#Events} & \textbf{\#Shots} & \textbf{Input} \\
    \midrule
    T2AV & 128 & 879 & 2,115 & S \\
    I2AV & 115 & 807 & 1,989 & RI+S \\
    V2AV & 41 & 235 & 731 & RV+S \\
    \bottomrule
    \end{tabularx}
  \end{table}

As shown in Table~\ref{tab:task_coverage}, \name covers three long-form audio-visual generation tasks under a unified benchmarking framework. 
In text-to-audio-video (T2AV), models generate minute-scale audio-visual content from structured event scripts. 
In image-to-audio-video (I2AV), models generate long-form sequences conditioned on a reference image and an event script, requiring consistent preservation of subject appearance and scene attributes throughout the generation process. 
In video-to-audio-video (V2AV), models extend a reference video according to a continuation script while preserving style consistency, subject continuity, temporal coherence, and audio-visual alignment.

\begin{figure}[t]
  \centering
  \includegraphics[width=\columnwidth]{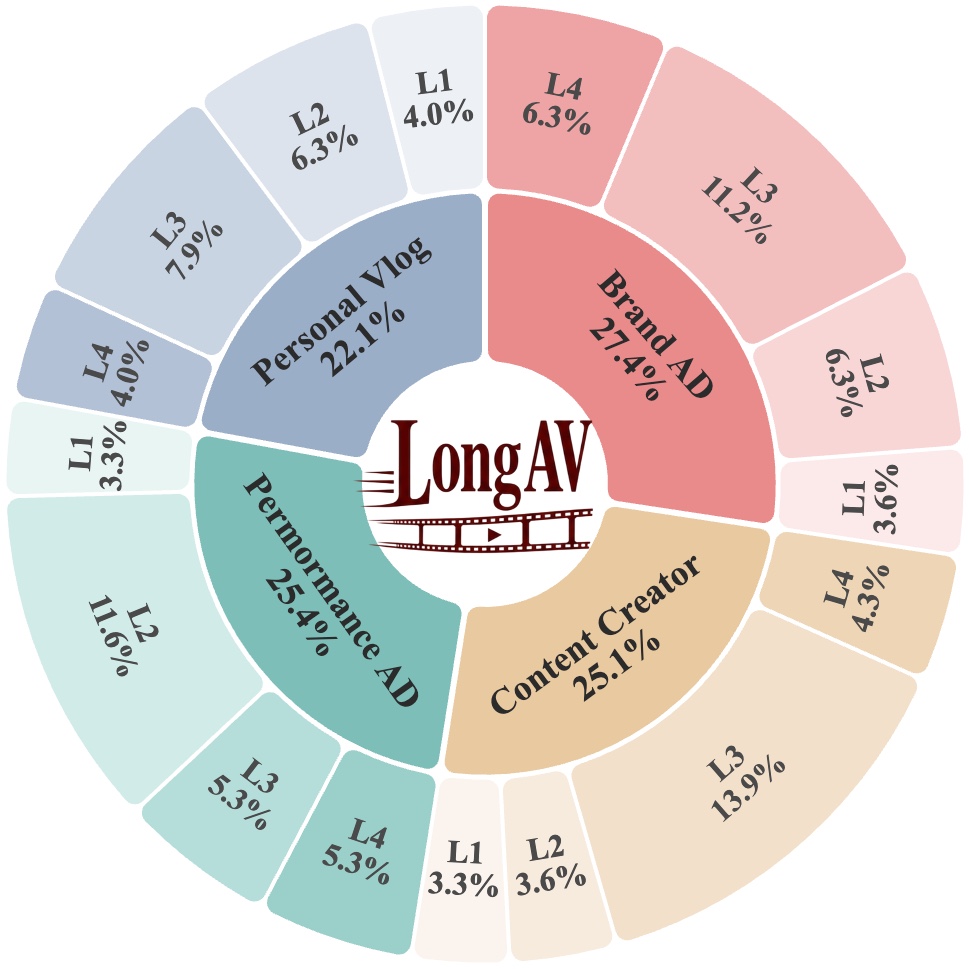}
  \caption{\textbf{Scenario and difficulty distribution in \name}. The
  benchmark spans four application scenarios and multiple complexity
  levels (L1--L4), supporting analysis by both content domain and generation
  difficulty.}
  \label{fig:taxonomy_sunburst}
  \Description{A sunburst chart showing the distribution of
  LongAV-Bench samples across application scenarios and difficulty
  levels.}
\end{figure}
This formulation treats conditioning modality as a unified evaluation dimension rather than separating tasks into independent benchmarks.
Accordingly, models are grouped according to the input interfaces they support, enabling unified evaluation across T2AV, I2AV, and V2AV settings.

\begin{figure*}[t] 
  \centering
   \includegraphics[width=\textwidth,height=0.88\textheight,keepaspectratio]{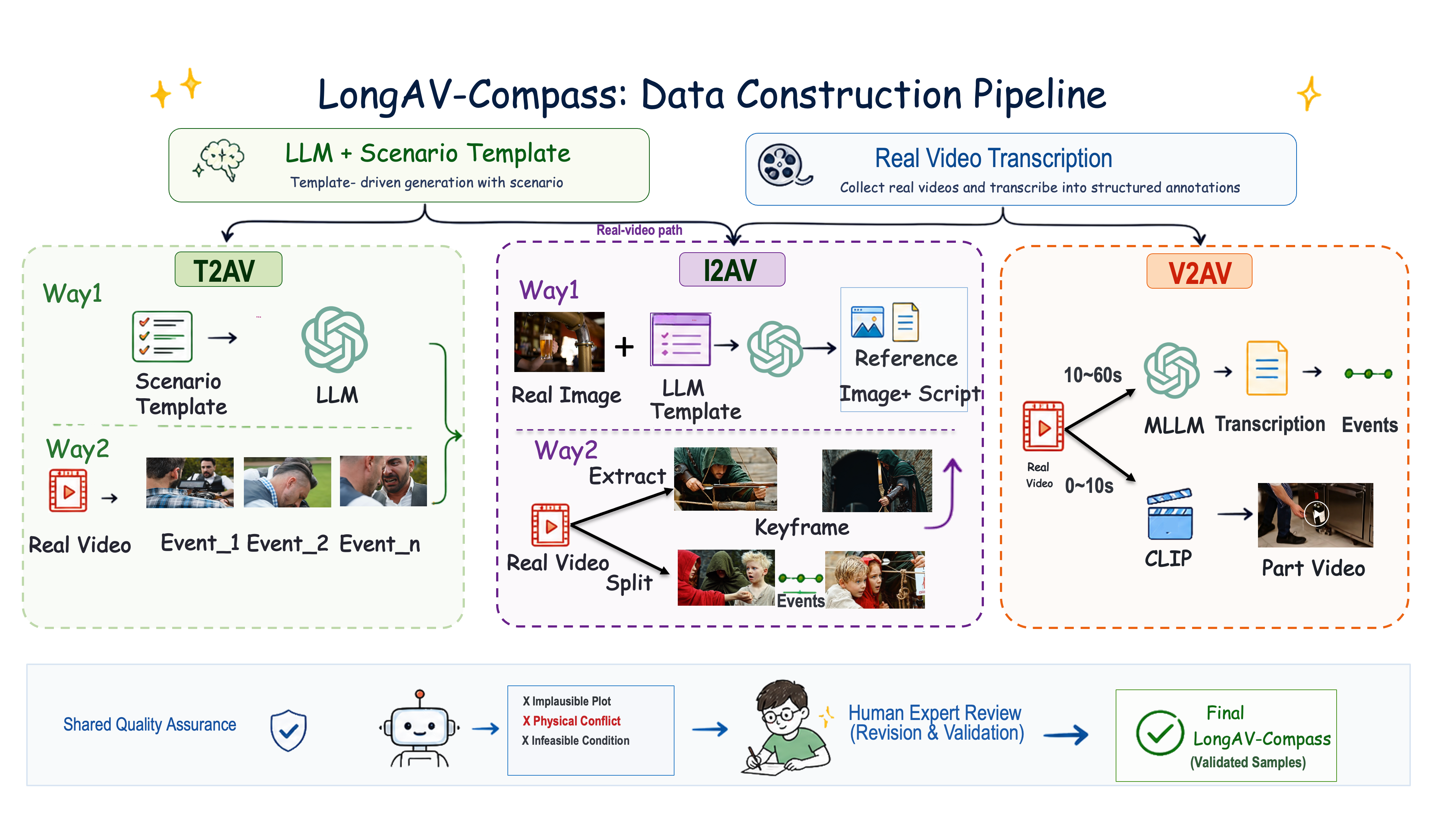}
  \caption{\textbf{Data construction pipeline of \name.}
  \name builds its benchmark data for three task types: T2AV, I2AV, and V2AV.
  T2AV and I2AV cases are obtained through two complementary routes: scenario-
  template-based LLM generation and real-video-based transcription or
  adaptation. V2AV cases are constructed from real videos by extracting
  reference clips and generating continuation scripts. After task-specific
  construction, all cases are converted into a shared event-level annotation
  format and filtered through dual quality control with MLLM review and human
  validation.}
  \label{fig:data_construction_pipeline}
  \Description{A full-page pipeline diagram showing task-specific data
  construction for T2AV, I2AV, and V2AV, followed by shared quality
  assurance and human expert review.}
\end{figure*}

\subsection{Taxonomy and Benchmark Scope}

\name is organized by a two-dimensional taxonomy defined over \emph{application scenario} and \emph{generation complexity}. 
The scenario axis covers four settings: \textit{Vlog}, \textit{Content-Creator}, \textit{Performance Ads}, and \textit{Brand Ads}. 
Here, \textit{Content-Creator} denotes structured creator-oriented content, such as comic drama generation and AI short dramas; \textit{Performance Ads} refers to platform-oriented promotional content, such as e-commerce or conversion-driven campaigns; and \textit{Brand Ads} targets large-scale brand marketing. 
This scenario design prevents the benchmark from being dominated by a single narrative genre and enables evaluation across both informal user-generated content and highly structured commercial generation settings.
The complexity axis contains four levels. \emph{L1} focuses on multiple entities or simple short-range interactions; \emph{L2} introduces multi-event structures and cross-event transitions; \emph{L3} emphasizes multi-actor interactions, role consistency, and longer-range dependency tracking; and \emph{L4} targets causal chains, physical plausibility, and more demanding story closure. 
Together, these axes make generation difficulty explicit and allow model performance to be analyzed as a function of structural complexity rather than only through aggregate scores.
Figure~\ref{fig:taxonomy_sunburst} visualizes the resulting distribution across application scenarios and difficulty levels, showing that \name supports analysis along both content-domain and generation-complexity axes.
Prompt detail is treated as an orthogonal variable rather than being tied to a specific scenario type. 
Each scenario includes short, medium, and long instructions. 
Short prompts test whether a model can expand an underspecified request into a coherent minute-long sequence, whereas long prompts stress fine-grained controllability and script following.

\subsection{Data Construction}

\paragraph{T2AV Task.}
The T2AV split contains $128$ cases constructed through a two-track pipeline. 
Approximately $60$\% of the scripts are derived from real videos with open or permissive licenses, while the remaining $40$\% are generated from scenario-by-complexity templates with LLM assistance. 
For the real-video track, we collect $50$--$90$ second videos from sources such as YouTube videos released under Creative Commons licenses, FineVideo, Pexels, and Pixabay, and use Gemini 3.1 Pro~\cite{gemini31pro} to convert them into structured long-form scripts. 
For the template-based track, human designers first specify scenario templates, complexity targets, and prompt-detail levels, after which Gemini 3.1 Pro generates paired global descriptions and event-level sequences. 
Both tracks are further filtered through human review to ensure physical plausibility, generation feasibility, and diagnostic value. Figure~\ref{fig:data_construction_pipeline} summarizes the task-specific construction pipelines.

\paragraph{I2AV Task.}
The I2AV split contains $115$ reference-image cases. 
Images are collected from permissively licensed repositories, including Pixabay, Burst, StockSnap, and Pexels, with balanced coverage across the same scenario taxonomy. 
For each image, Gemini 3.1 Pro generates a long-form audio-visual description in two aligned formats: a global narrative and a sequence of timed events. 
Human reviewers then verify whether the description is faithful to the visible image content, whether the inferred action sequence is physically plausible, and whether the case is suitable for minute-long generation.

\paragraph{V2AV Task.}
The V2AV split contains $41$ reference-video continuation cases. 
Each case consists of a $10$--$15$ second reference clip and a textual continuation script for the remaining $45$--$50$ seconds. 
Reference clips are collected from open-license sources or reused from the real-video track when they provide a clean continuation boundary. 
Gemini 3.1 Pro proposes the continuation script, and human reviewers validate whether the continuation is natural, generation-feasible, and informative for evaluating long-range transition quality.

\subsection{Unified Annotation Format}

Each case in \name is annotated with two coupled representations: a global description and an event sequence. 
The global description summarizes the overall intent, narrative structure, and expected audio-visual outcome of the minute-long generation, and serves as the primary conditioning input for model generation. 
The event sequence decomposes the case into temporally aligned sub-events and provides structured support for event-level evaluation and fine-grained diagnosis. 
Each event specifies a temporal span, an action summary, a completion criterion, key visual elements, and the expected audio content. 
This dual representation enables both high-level semantic assessment and event-aligned diagnostics. 
In addition, we annotate identity constraints, physical constraints, and narrative dependencies to specify which elements should remain stable or logically consistent across the generated output.

Task-specific fields are added when required by the conditioning modality. 
I2AV cases include a reference image, a subject description, and identity constraints that define appearance anchors. 
V2AV cases include a reference video, a reference-video description, and a continuation description. 
This unified yet task-aware schema enables comparison across T2AV, I2AV, and V2AV while preserving their distinct conditioning requirements.

\subsection{Video Metrics}

To systematically evaluate long-form video generation, \name defines six shared video metrics spanning event fulfillment, segment-level quality, long-range continuity, transition stability, holistic presentation, and text-video alignment. 
Together, these metrics provide complementary views of generation quality at the event, segment, and full-video levels.

\noindent\textbf{Event fulfillment ($\mathbf{V}_{\mathrm{QA}}$).} 
For each event, we construct content-oriented questions from the event annotation and use an MLLM to verify whether the required subjects, actions, and visual details are correctly reflected in the generated video. The resulting event-completion score is normalized to the range of $0$--$1$.

\noindent\textbf{Visual quality (VQ).} 
We evaluate each event segment with an MLLM along four local visual dimensions: motion naturalness, subject integrity, artifact control, and visual fidelity. The final VQ score is reported on a $1$--$5$ scale.

\noindent\textbf{Long-form continuity (Cont.).} 
This metric measures whether the generated video remains coherent over the full temporal horizon. We extract low-frame-rate previews from the complete video and evaluate them together with the global description and event sequence. A multimodal evaluator scores story continuity, subject consistency, scene coherence, and temporal progression on a $1$--$5$ scale, and the final Cont. score is computed as a weighted average.

\noindent\textbf{Transition stability (Trans.).} 
We evaluate event boundaries by checking for black frames, flickering, repetition, freezing, and abrupt visual discontinuities, and combine these signals with MLLM-based judgments of boundary-level breaks. The Trans. score is reported on a $1$--$5$ scale.

\noindent\textbf{Holistic presentation (Hol.).} 
We evaluate the complete video as a finished work, considering style consistency, visual appeal, commercial completeness, and overall watchability. Unlike continuity, which focuses on temporal coherence, Hol. captures the overall presentation quality and perceived completeness of the generated video. The Hol. score is reported on a $1$--$5$ scale.

\noindent\textbf{Text-video alignment (TVAlign).} 
We measure whether the full video remains semantically aligned with the global description and event sequence. Specifically, TVAlign is computed using CLIP embedding similarity\cite{radford2021learning} between the textual description and sampled video frames, and is reported as a $0$--$1$ score.

\subsection{Audio Metrics}

To evaluate long-form audio generation and cross-modal synchronization, \name defines three audio metrics covering temporal alignment, event-level audio quality, and long-range soundtrack coherence. 
These metrics are applied to models with native audio generation capability, while models without an audio track are still evaluated under the shared video metrics and marked as N/A for audio evaluation.

\noindent\textbf{Audio-video synchronization (AVS).} 
We measure whether speech, sounds, music changes, and sound effects are temporally aligned with the corresponding visible actions, scene transitions, and edits. The AVS score is reported on a $1$--$5$ scale.

\noindent\textbf{Audio quality (AudQ).} 
We evaluate the realism and event-level appropriateness of the generated audio with respect to the event text and audio expectation. This includes whether sound sources are plausible, whether the audio content matches the visual scene, and whether obvious artifacts are absent. The AudQ score is reported on a $1$--$5$ scale.

\noindent\textbf{Long-audio coherence (AudL).} 
We evaluate whether the full soundtrack remains continuous and stable over the complete video, without abrupt silence, unnatural repetition, volume jumps, or disruptive transitions. The AudL score is reported on a $1$--$5$ scale.

\subsection{Task-Specific Metrics}

For I2AV, we define two task-specific metrics to measure reference-image preservation. 
First-frame image anchoring ($\mathrm{IV}_1$) evaluates whether the opening frame of the generated video preserves the subject appearance and scene attributes specified by the reference image. 
Image alignment (ImgAlign) further measures whether this reference-image consistency is maintained over time. 
Specifically, we compute CLIP image-image similarity between the reference image and sampled frames from each generated event segment. 
The event-level ImgAlign score is obtained by averaging the similarities over sampled frames, and the final video-level score is computed by averaging event-level scores across the full video.

\section{Experiments}
\label{sec:experiments}

\subsection{Experimental Settings}

\begin{table*}[t]
  \caption{\textbf{Main results on T2AV task.} We report event-level fulfillment and quality, long-form consistency, global presentation, text-video alignment, and audio diagnostics. The highest score in each dimension is \textbf{boldfaced} and highlighted in \colorbox{softgreen}{green}.}
  \label{tab:t2av_main_results}
  \scriptsize
  \centering
  \resizebox{\textwidth}{!}{%
  \begin{tabular}{lcccccccccc}
\toprule
\multirow{2}{*}{\textbf{Model}} &
\multirow{2}{*}{\textbf{Aud.}} &
\multicolumn{2}{c}{\textbf{Event}} &
\multicolumn{2}{c}{\textbf{Consistency}} &
\textbf{Global Pres.} &
\textbf{Text Align.} &
\multicolumn{3}{c}{\textbf{Audio}} \\
\cmidrule(lr){3-4}\cmidrule(lr){5-6}\cmidrule(lr){9-11}

& &
$\mathbf{V}_{\mathrm{QA}}$ & \textbf{VQ} &
\textbf{Cont.} & \textbf{Trans} &
\textbf{Hol.} & \textbf{TVAlign} &
\textbf{AVS} & \textbf{AudQ} & \textbf{AudL} \\
\midrule

\rowcolor{softblue}
\multicolumn{11}{c}{\textbf{Proprietary Models}} \\

Seedance 2.0 & Yes & 0.9023 & \cellcolor{softgreen}\textbf{3.7116} & 4.2649 & 4.0065 & \cellcolor{softgreen}\textbf{4.1128} & 0.6183 &
\cellcolor{softgreen}\textbf{3.6038} & \cellcolor{softgreen}\textbf{3.7875} & \cellcolor{softgreen}\textbf{4.1845} \\

Kling 3.0 & Yes & \cellcolor{softgreen}\textbf{0.9274} & 3.3893 & \cellcolor{softgreen}\textbf{4.4139} & 3.8502 & 3.8542 & 0.6185 &
3.4922 & 3.6049 & 3.7713 \\

Veo 3.1 & Yes & 0.7784 & 2.8961 & 3.1348 & 4.0032 & 3.5759 & 0.6142 &
3.3490 & 3.2387 & 3.6931 \\

\rowcolor{softpink}
\multicolumn{11}{c}{\textbf{Open-Source Models}} \\

LTX 2.3 & Yes & 0.7321 & 2.2880 & 3.2888 & 3.8829 & 3.0203 & \cellcolor{softgreen}\textbf{0.6205} &
2.7278 & 2.5017 & 2.9313 \\

Longcat & Yes & 0.5870 & 2.0310 & 2.0735 & 3.8907 & 2.5176 & 0.6148 &
-- & -- & -- \\

Wan2.2-I2V-A14B & No & 0.5994 & 2.0046 & 2.2576 & 3.5747 & 2.6794 & 0.6123 &
N/A & N/A & N/A \\

HunyuanVideo 1.5-I2V & No & 0.5772 & 1.9790 & 1.9199 & \cellcolor{softgreen}\textbf{4.1598} & 2.4880 & 0.6165 &
N/A & N/A & N/A \\

Helios (14B) & No & 0.5013 & 1.9370 & 1.8294 & 3.3490 & 2.5912 & 0.6152 &
N/A & N/A & N/A \\

Open-Sora & No & 0.2476 & 1.3854 & 1.4947 & 3.6418 & 1.5676 & 0.6161 &
N/A & N/A & N/A \\

davinci-magihuman & Yes & 0.4583 & 1.7100 & 1.9306 & 2.7602 & 2.3535 & 0.6116 &
2.8063 & 2.4622 & 2.9856 \\

\rowcolor{softyellow}
\multicolumn{11}{c}{\textbf{Agent-Based Models}} \\

VideoDirectorGPT & No & 0.5205 & 2.0990 & 1.8172 & 3.3830 & 2.4549 & 0.6155 &
N/A & N/A & N/A \\

  \bottomrule
  \end{tabular}}
\end{table*}

\begin{table*}[t]
  \caption{\textbf{Main results on I2AV task.} In addition to shared video and audio diagnostics, we report image alignment through first-frame anchoring and CLIP-based event-level image-video alignment. The highest score in each dimension is \textbf{boldfaced} and highlighted in \colorbox{softgreen}{green}.}
  \label{tab:i2av_main_results}
  \scriptsize
  \centering

  \resizebox{\textwidth}{!}{%
  \begin{tabular}{lcccccccccccc}
\toprule
\multirow{2}{*}{\textbf{Model}} &
\multirow{2}{*}{\textbf{Aud.}} &
\multicolumn{2}{c}{\textbf{Event}} &
\multicolumn{2}{c}{\textbf{Consistency}} &
\textbf{Global Pres.} &
\textbf{Text Align.} &
\multicolumn{2}{c}{\textbf{Image Align.}} &
\multicolumn{3}{c}{\textbf{Audio}} \\
\cmidrule(lr){3-4}\cmidrule(lr){5-6}\cmidrule(lr){9-10}\cmidrule(lr){11-13}

& &
$\mathbf{V}_{\mathrm{QA}}$ & \textbf{VQ} &
\textbf{Cont.} & \textbf{Trans} &
\textbf{Hol.} & \textbf{TVAlign} &
$\mathbf{IV}_1$ & \textbf{ImgAlign} &
\textbf{AVS} & \textbf{AudQ} & \textbf{AudL} \\
\midrule

\rowcolor{softblue}
\multicolumn{13}{c}{\textbf{Proprietary Models}} \\

Seedance 2.0 & Yes & \cellcolor{softgreen}\textbf{0.9204} & \cellcolor{softgreen}\textbf{3.7651} & \cellcolor{softgreen}\textbf{4.9182} & 3.9625 & \cellcolor{softgreen}\textbf{3.8864} & 0.6145 &
0.9622 & 0.9027 & \cellcolor{softgreen}\textbf{3.5669} & \cellcolor{softgreen}\textbf{3.9113} & \cellcolor{softgreen}\textbf{4.2290} \\

Kling 3.0 & Yes & 0.8939 & 3.2760 & 4.1244 & 4.0668 & 3.8526 & 0.6182 &
\cellcolor{softgreen}\textbf{0.9960} & 0.8877 & 3.5081 & 3.8032 & 4.0164 \\

Veo 3.1 & Yes & 0.8211 & 2.9266 & 3.8183 & 4.1414 & 3.6463 & 0.6156 &
0.9685 & 0.9051 & 3.3514 & 3.4484 & 4.1221 \\

\rowcolor{softpink}
\multicolumn{13}{c}{\textbf{Open-Source Models}} \\

Wan2.2-I2V-A14B & No & 0.6832 & 2.2526 & 2.5340 & 4.0762 & 2.7926 & 0.6120 &
0.9667 & 0.8999 & N/A & N/A & N/A \\

Longcat & Yes & 0.5954 & 2.0632 & 2.1277 & 4.1625 & 2.4574 & 0.6155 &
0.9227 & 0.9006 & -- & -- & -- \\

LTX 2.3 & Yes & 0.6967 & 2.1121 & 3.1441 & 3.8649 & 2.7473 & \cellcolor{softgreen}\textbf{0.6191} &
0.9122 & 0.8728 & 2.7017 & 2.5322 & 2.7940 \\

HunyuanVideo 1.5-I2V & No & 0.5934 & 1.9425 & 1.8267 & \cellcolor{softgreen}\textbf{4.1868} & 2.3807 & 0.6153 & 0.9351 & 0.9160 & N/A & N/A & N/A \\

Helios (14B) & No & 0.4620 & 1.8006 & 1.8133 & 3.4678 & 2.3750 & 0.6125 &
0.9186 & 0.9202 & N/A & N/A & N/A \\

davinci-magihuman & Yes & 0.4860 & 1.6519 & 1.6734 & 3.1634 & 2.0691 & 0.6131 &
0.9223 & 0.9050 & 2.7172 & 2.4271 & 2.9160 \\

Open-Sora & No & 0.3009 & 1.4669 & 1.3032 & 3.7476 & 1.5678 & 0.6153 &
0.9133 & 0.9184 & N/A & N/A & N/A \\

\rowcolor{softyellow}
\multicolumn{13}{c}{\textbf{Agent-Based Models}} \\

VideoDirectorGPT & No & 0.1976 & 1.5073 & 1.0000 & 3.3935 & 1.7378 & 0.6033 &
0.9303 & \cellcolor{softgreen}\textbf{0.9640} & N/A & N/A & N/A \\
  \bottomrule
  \end{tabular}}
\end{table*}

\begin{table*}[t]
    \caption{\textbf{Main results on V2AV task.} We report event-level
  fulfillment and quality, long-form consistency, global presentation, text-
  video alignment, and audio diagnostics for video continuation. The highest
  score in each dimension is \textbf{boldfaced} and highlighted in
  \colorbox{softgreen}{green}.}
    \label{tab:v2av_main_results}
    \scriptsize
    \centering

    \resizebox{\textwidth}{!}{%
    \begin{tabular}{lcccccccccc}
  \toprule
  \multirow{2}{*}{\textbf{Model}} &
  \multirow{2}{*}{\textbf{Aud.}} &
  \multicolumn{2}{c}{\textbf{Event}} &
  \multicolumn{2}{c}{\textbf{Consistency}} &
  \textbf{Global Pres.} &
  \textbf{Text Align.} &
  \multicolumn{3}{c}{\textbf{Audio}} \\
  \cmidrule(lr){3-4}\cmidrule(lr){5-6}\cmidrule(lr){9-11}

  & &
  $\mathbf{V}_{\mathrm{QA}}$ & \textbf{VQ} &
  \textbf{Cont.} & \textbf{Trans} &
  \textbf{Hol.} & \textbf{TVAlign} &
  \textbf{AVS} & \textbf{AudQ} & \textbf{AudL} \\
  \midrule

  \rowcolor{softblue}
  \multicolumn{11}{c}{\textbf{Proprietary Models}} \\

  Seedance 2.0 & Yes & \cellcolor{softgreen}\textbf{0.8753} &
  \cellcolor{softgreen}\textbf{3.8336} & \cellcolor{softgreen}\textbf{4.7636} &
  3.9267 & \cellcolor{softgreen}\textbf{4.1705} &
  \cellcolor{softgreen}\textbf{0.9727} &
  \cellcolor{softgreen}\textbf{3.7591} & \cellcolor{softgreen}\textbf{4.4357} &
  \cellcolor{softgreen}\textbf{4.3129} \\

  Veo 3.1 & Yes & 0.8055 & 3.0869 & 1.8425 & 2.2815 & 3.3625 & 0.7100 &
  3.4939 & 3.9485 & 3.2897 \\

  \rowcolor{softpink}
  \multicolumn{11}{c}{\textbf{Open-Source Models}} \\

  Helios (14B) & No & 0.4818 & 1.8197 & 2.0324 & 3.9222 & 2.2206 & 0.5191 &
  N/A & N/A & N/A \\

  Longcat & No & 0.5031 & 1.8937 & 1.5809 & \cellcolor{softgreen}\textbf{3.9848}
  & 2.1691 & 0.3706 &
  N/A & N/A & N/A \\

  Helios-Distilled & No & 0.3559 & 1.6365 & 1.4515 & 3.8092 & 1.7941 & 0.3529 &
  N/A & N/A & N/A \\

    \bottomrule
    \end{tabular}}
  \end{table*}
\paragraph{Implementation Details.}
All local model inference, video post-processing, and metric computation are conducted on servers equipped with NVIDIA H200 GPUs. 
For models accessed through commercial services, we use their official generation APIs when available, or official web interfaces otherwise. 
To ensure fair comparison, all submitted prompts are derived from the same benchmark annotations, with only format-level adaptations made to match each model's native input interface. 
Unless otherwise specified, we preserve the default generation configuration of each model. 
Detailed generation prompts, model-specific adaptations, output processing procedures, and evaluation rubrics are provided in Appendix~\ref{app:eval_framework} and Appendix~\ref{app:generation_protocol}.
  
\paragraph{Evaluated Models.}
We evaluate $11$ representative video generation systems and group them into three categories: \textit{proprietary models}, \textit{open-source models}, and \textit{agent-based models}. 
The proprietary models include Seedance 2.0~\cite{seedance2026seedance}, Kling 3.0~\cite{team2025kling}, and Veo 3.1. 
The open-source models include LTX 2.3~\cite{hacohen2024ltx}, LongCat~\cite{longcat2025longcat}, Wan2.2-I2V-A14B~\cite{wan2025wan}, HunyuanVideo 1.5-I2V~\cite{wu2025hunyuanvideo15}, Helios (14B)~\cite{yuan2026helios}, Open-Sora~\cite{zheng2025opensora2}, and daVinci-MagiHuman~\cite{siigair2026davincimagihuman}. 
We also include VideoDirectorGPT~\cite{lin2023videodirectorgpt} as an agent-based baseline.

\paragraph{Evaluation Protocol and Fairness Controls.}
For all tasks, the target output duration is at least $60$ seconds and typically falls within the $60$--$120$ second range. 
We preserve each system's native generation configuration whenever possible, including its default generation pipeline, resolution, temporal sampling strategy, and audio interface. 
When a model requires task-specific prompt syntax or multi-stage orchestration, we convert the benchmark input into the closest native format while preserving event order, conditioning semantics, and audio expectations. 
Models without native audio are evaluated under the shared video-only protocol, while models with native audio are additionally included in audio-visual evaluation.

\subsection{Evaluation Framework Overview}
\name adopts a unified diagnostic evaluation framework for minute-long audio-visual generation across T2AV, I2AV, and V2AV. 
As shown in Fig.~\ref{fig:overview}, the framework combines event-aligned segment evaluation, full-video assessment, and task-specific reference checks. 
Rather than collapsing all signals into a single score, we report complementary diagnostic dimensions that separately measure event fulfillment, segment-level generation quality, long-range consistency, global presentation, semantic alignment, audio quality, and task-specific image anchoring or video continuation behavior.

\paragraph{Task-Aligned Segmentation.}
Failures in long-form generation often emerge around event boundaries or accumulate across temporally distant segments. 
Therefore, we evaluate outputs using event-aligned segments rather than fixed temporal windows. 
For T2AV, segments follow the event structure of the input script. 
For I2AV, the same event-aligned segmentation is retained and paired with the reference image to support image-alignment diagnostics. 
For V2AV, segmentation is anchored at the reference-video boundary and follows the annotated continuation events. 
We also extract boundary clips around adjacent events to assess transition stability.

\paragraph{Reporting Protocol.}
We report all metrics as diagnostic dimensions in task-specific result tables. 
T2AV uses the shared video and audio metrics; I2AV further includes image-conditioned metrics such as $\mathrm{IV}_1$ and ImgAlign; and V2AV includes video-continuation metrics that assess reference consistency and continuation quality. 
This protocol keeps evaluation comparable across tasks while preserving the task-specific signals needed to diagnose image anchoring and video continuation failures.

\begin{figure}[t]
  \centering
  \includegraphics[width=\columnwidth]{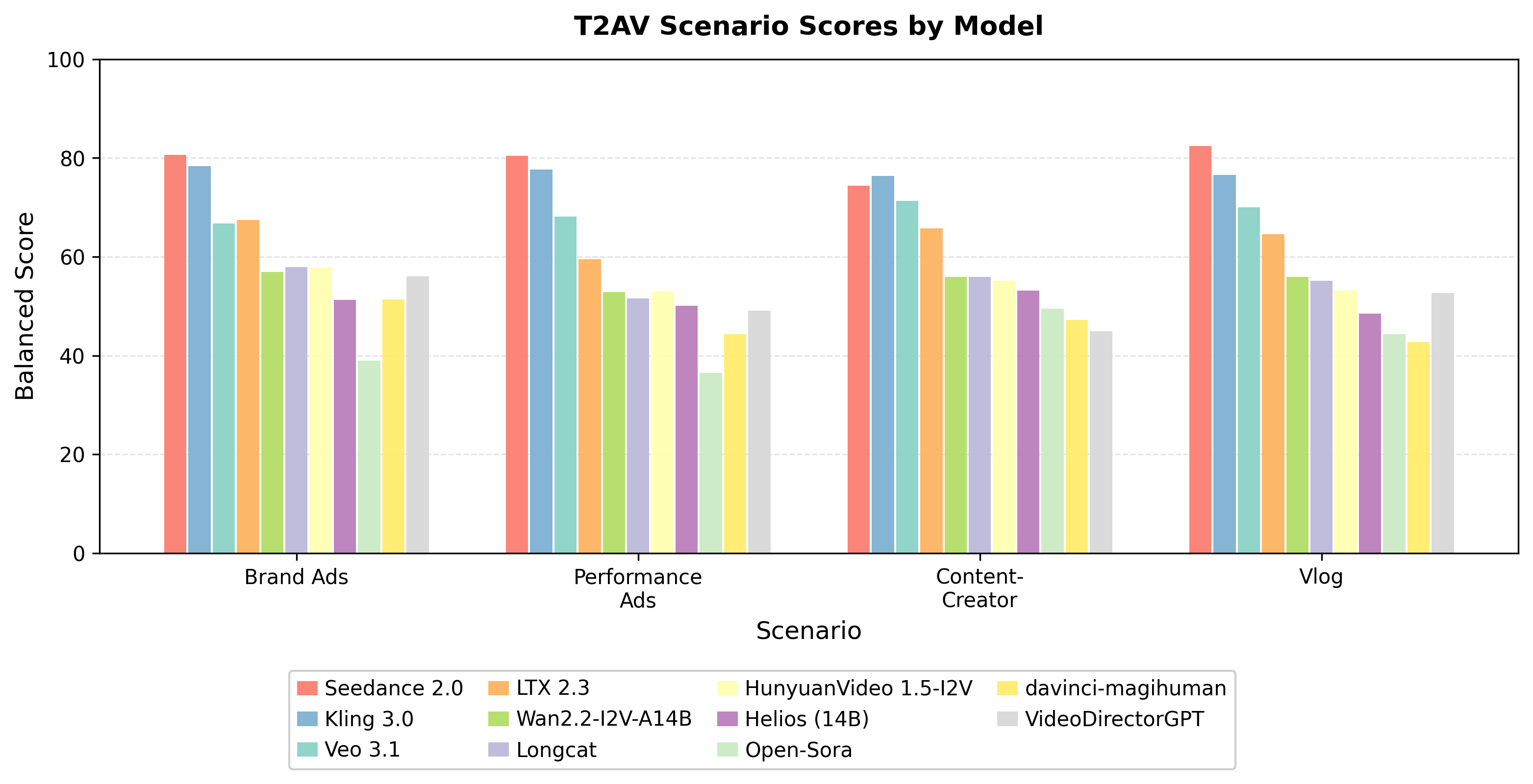}
  \caption{\textbf{Scenario-level balanced scores on T2AV task.} For each scenario, each bar reports the mean balanced score of one model over all available samples in that scenario.}

  \label{fig:t2av_scenario_model_scores}
\end{figure}

\begin{figure}[t]
  \centering
  \includegraphics[width=\columnwidth]{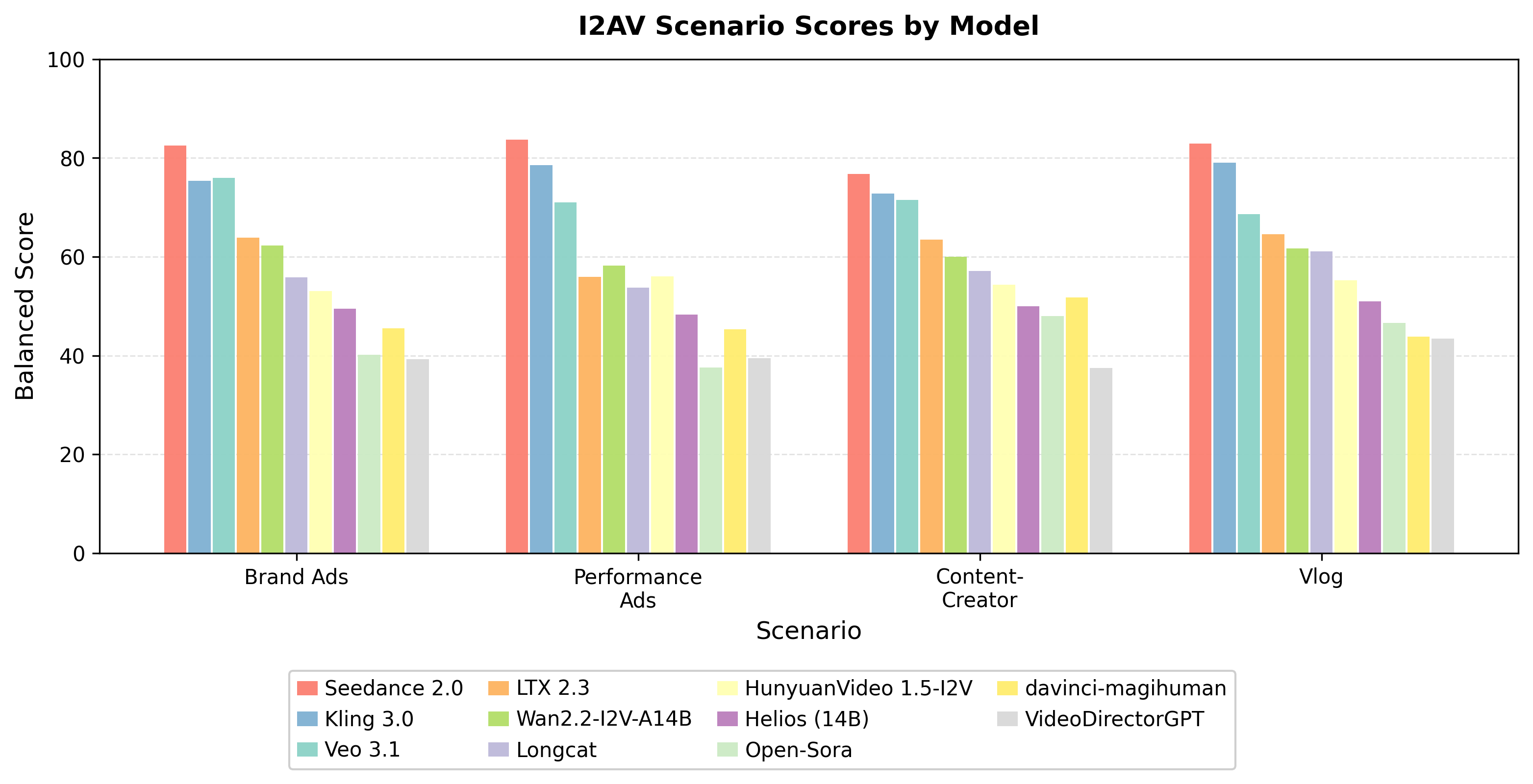}
  \caption{\textbf{Scenario-level balanced scores on I2AV task.} For each scenario, each bar reports the mean balanced score of one model over all available samples in that scenario.}
  \label{fig:i2av_scenario_model_scores}
\end{figure}


\subsection{Main Results}

Tables~\ref{tab:t2av_main_results}--\ref{tab:v2av_main_results} show that current long-form audio-visual generation systems cannot be adequately characterized by a single overall score. 
Strong performance requires the joint ability to execute annotated events, maintain long-range temporal structure, produce visually coherent outputs, and generate stable audio. 
From this diagnostic perspective, Seedance 2.0 is the most consistent model: it leads or remains near the top on T2AV and I2AV, and shows particularly strong performance on V2AV. 
Kling 3.0 and Veo 3.1 are also competitive on several dimensions, suggesting that model strengths vary substantially across evaluation axes.

\noindent\textbf{T2AV Task.} 
Under script-only conditioning, leading models exhibit different strengths. Kling 3.0 achieves the highest event-fulfillment score ($\mathrm{V}_{\mathrm{QA}}=0.9274$) and the strongest long-form continuity (Cont.=$4.4139$), indicating reliable coverage of the requested event sequence. 
In contrast, Seedance 2.0 obtains the best visual quality (VQ=$3.7116$), holistic presentation (Hol.=$4.1128$), and all three audio scores, suggesting stronger overall audio-visual generation quality. 
The results also reveal that strong performance on individual proxy metrics does not necessarily translate into successful long-form generation. 
For example, LTX 2.3 achieves the highest TVAlign score, and HunyuanVideo 1.5-I2V obtains the highest transition score, yet both remain substantially behind the leading proprietary models in event fulfillment, visual quality, and holistic presentation. 
This suggests that embedding-level semantic alignment or smooth local transitions alone are insufficient for minute-long script realization.

\noindent\textbf{I2AV Task.} 
As shown in Table~\ref{tab:i2av_main_results}, Seedance 2.0 remains the strongest overall I2AV model, while Kling 3.0 leads in first-frame anchoring and Veo 3.1 performs well on transition stability. 
Notably, several models achieve high reference-image similarity according to $\mathrm{IV}_1$ and ImgAlign despite much lower event-level and continuity scores. 
For example, VideoDirectorGPT obtains the highest ImgAlign score ($0.9640$) but performs poorly in event fulfillment and continuity, while Helios and Open-Sora also retain competitive image-similarity scores despite limited visual quality and holistic presentation. 
These results indicate that reference-image preservation is necessary but insufficient for I2AV: models must also infer plausible motion, organize event progression, and avoid temporal drift over the full generation duration.

\noindent\textbf{V2AV Task.} 
As shown in Table~\ref{tab:v2av_main_results}, Seedance 2.0 clearly dominates this split, leading in event fulfillment, visual quality, continuity, holistic presentation, text-video alignment, and audio quality. 
Veo 3.1 retains strong event fulfillment, but its substantially lower continuity and transition scores suggest that a semantically plausible continuation can still fail at the reference boundary or drift away from the reference video's temporal state. 
Among open-source models, LongCat and Helios obtain relatively high transition scores but low holistic and alignment scores, indicating that they can produce locally smooth continuations while failing to preserve higher-level narrative structure and conditioning fidelity.

Across the three tasks, two patterns are particularly informative. 
First, task-specific alignment metrics often saturate or vary within a narrow range, whereas event fulfillment, continuity, and holistic presentation provide more discriminative signals for long-form generation quality. 
Second, native audio support alone does not guarantee synchronized, coherent, or event-appropriate soundtracks, as audio-capable models still differ substantially in long-form audio quality. 
These findings highlight the importance of evaluating minute-scale audio-visual generation beyond appearance preservation or local smoothness, with particular attention to event completeness, narrative progression, cross-event transitions, and audio-visual synchronization.

\subsection{Analysis and Findings}

\begin{figure*}[t] 
  \centering
   \includegraphics[width=\textwidth,height=0.88\textheight,keepaspectratio]{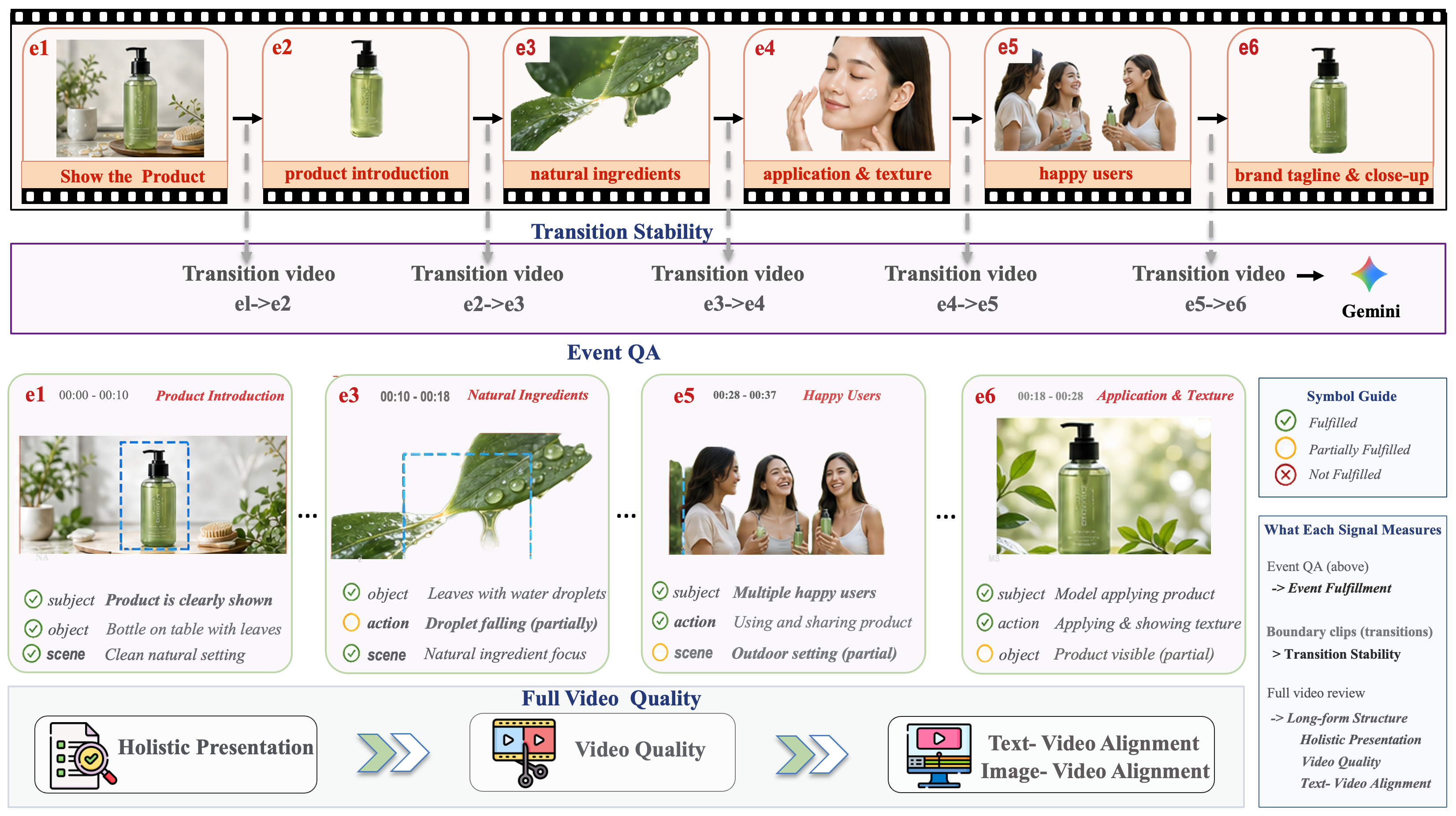}
 \caption{\textbf{Case study of event-aligned evaluation in \name.}
  Using a Brand Ads case as an example, the upper row decomposes the generated video into ordered events and boundary clips for transition-stability assessment. The middle row illustrates event-level QA for measuring event fulfillment, and the bottom row summarizes full-video quality signals, including holistic presentation, video quality, and text-/image-video
  alignment.}
  \label{fig:showcase}
  \Description{An example evaluation diagram showing event decomposition, transition clips, event-level QA checks, and full-video quality assessment for a Brand Ads generation case.}
\end{figure*}

\begin{figure}[t]
  \centering
  \includegraphics[width=0.8\columnwidth]{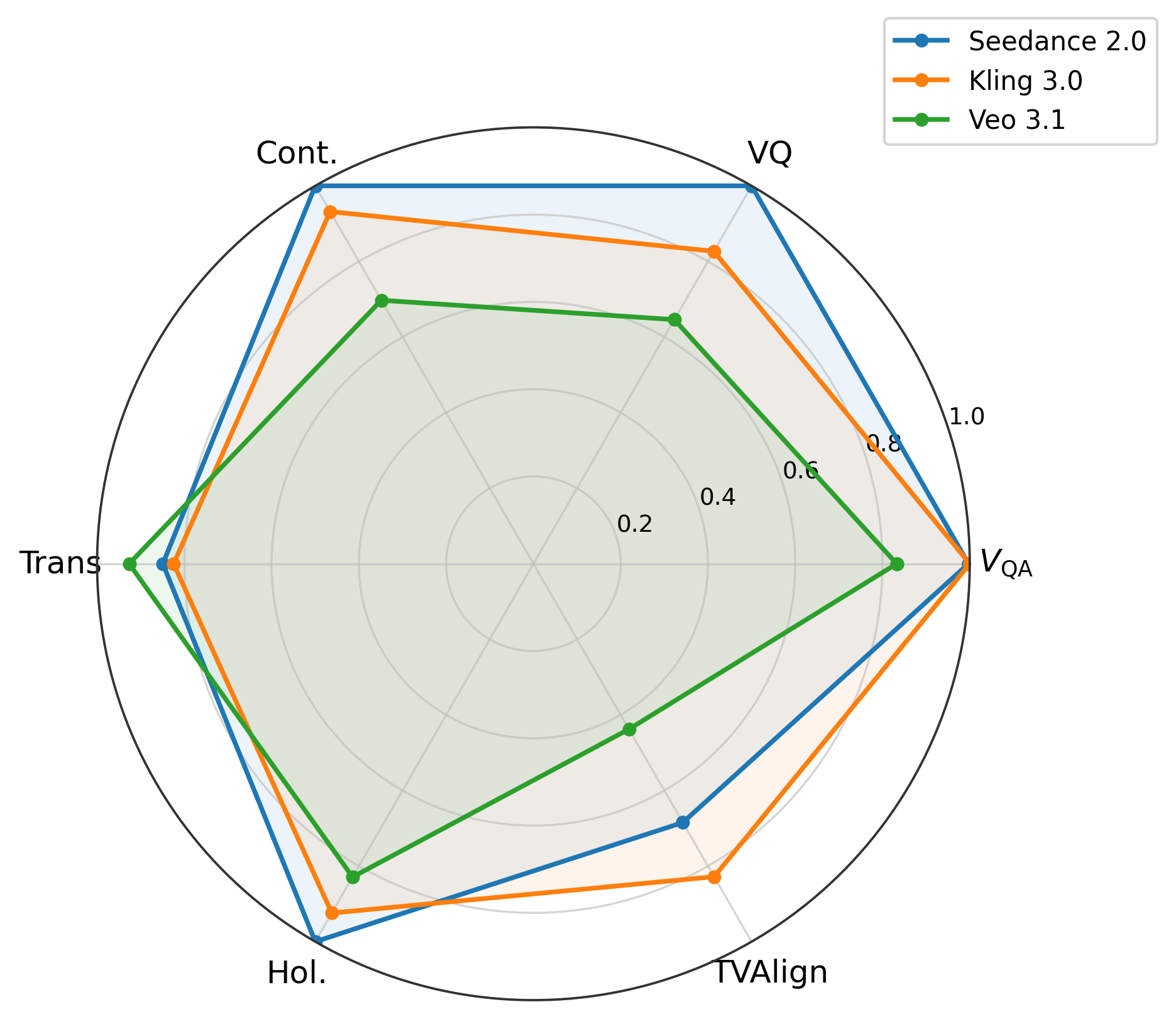}
  \caption{\textbf{Capability profiles of proprietary models.} Scores are min-max normalized per metric across the displayed models to highlight relative capability differences.}
  \label{fig:proprietary_capability_radar}
\end{figure}

\begin{figure}[t]
  \centering
  \includegraphics[width=0.8\columnwidth]{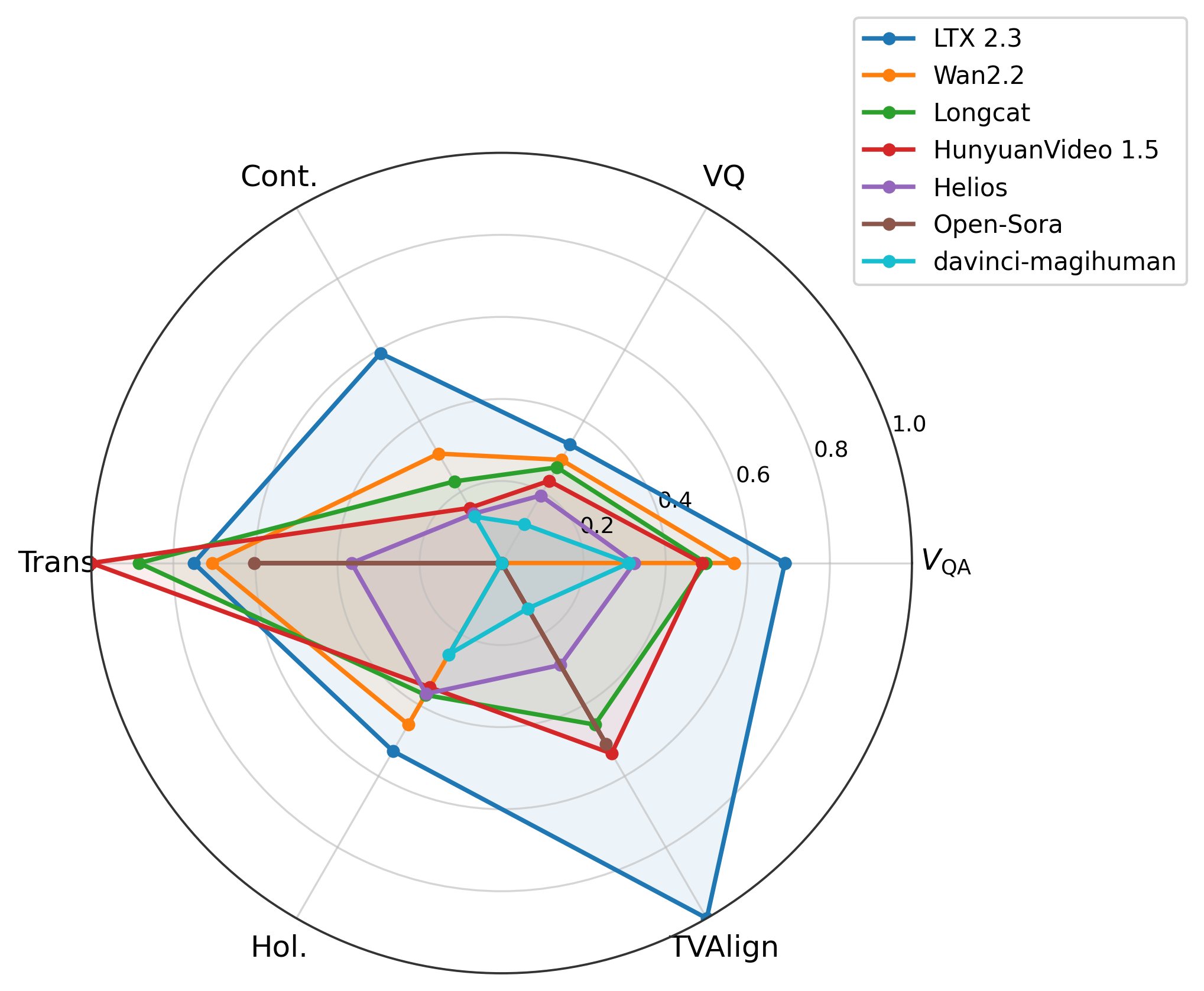}
  \caption{\textbf{Capability profiles of open-source models.} Scores are min-max normalized per metric across the displayed models to highlight relative capability differences.}
  \label{fig:opensource_capability_radar}
\end{figure}

Figure~\ref{fig:showcase} provides a concrete case study of the event-aligned evaluation process, illustrating how event fulfillment, transition stability, and full-video quality are assessed jointly. Figures~\ref{fig:proprietary_capability_radar}
  and~\ref{fig:opensource_capability_radar} compare normalized capability profiles within proprietary
  and open-source model groups, respectively.

\paragraph{Scenario-Level Behavior.}
We further analyze model performance across four application scenarios: Brand Ads, Performance Ads, Content-Creator, and Vlog. 
As shown in Figure~\ref{fig:t2av_scenario_model_scores} and~\ref{fig:i2av_scenario_model_scores}, proprietary models maintain clear advantages across all scenarios, indicating stronger event execution, cross-event organization, and holistic video completion. 
In contrast, open-source models and agent-based methods expose more pronounced weaknesses under scenario-specific requirements. 
Among individual systems, Seedance and Kling consistently rank near the top, while LTX 2.3 is the strongest open-source model and generally leads the open-source group across the four scenarios, although it still exhibits a persistent gap from proprietary systems. 
We also observe the largest score variance in the Performance Ads scenario, suggesting that this setting is particularly discriminative for separating model capabilities. 
Requirements such as product presentation, product consistency, multi-event explanation, and persuasive visual progression make Performance Ads a challenging testbed for current long-form audio-visual generation systems.


\begin{table}[t]
  \caption{\textbf{Per-difficulty analysis.} Each entry reports the average balanced score for a model family under one difficulty level.}
  \label{tab:complexity_analysis}
  \small
  \centering
  \begin{tabular}{lcccc}
  \toprule
  \textbf{Family} & \textbf{L1} & \textbf{L2} & \textbf{L3} & \textbf{L4} \\
  \midrule
  Proprietary Models & 70.6 & 75.2 & 74.5 & 73.9 \\
  Open-Source Models & 57.9 & 52.9 & 52.8 & 51.4 \\
  Agent-Based Models & 47.3 & 47.4 & 43.2 & 41.2 \\
  \bottomrule
  \end{tabular}
\end{table}

\begin{figure}[t]
  \centering
  \includegraphics[width=0.8\columnwidth]{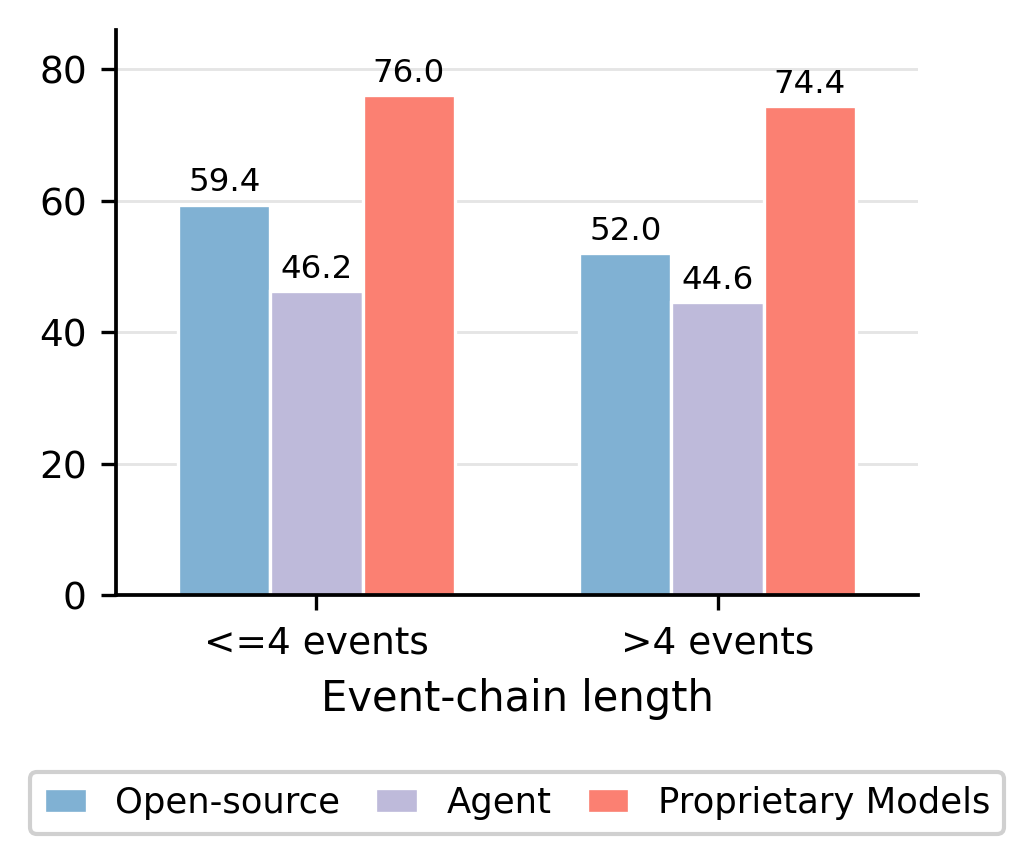}
  \caption{\textbf{Event-count analysis.} Samples are grouped into short event chains ($\leq$4 events) and longer event chains ($>$4 events), and each bar reports the average balanced score for one model family.}
  \label{fig:event_count_family_bars}
\end{figure}

\paragraph{Difficulty and Event-Count Effects.}
Table~\ref{tab:complexity_analysis} summarizes performance across progressively harder benchmark slices, revealing how model quality changes as generation requires longer causal structure and more complex multi-actor coordination.
Difficulty levels separate model families more clearly. 
When T2AV and I2AV are combined, commercial models remain relatively stable across the four difficulty levels, with composite scores ranging from $75.0$ to $73.9$. Open-source models are consistently lower, decreasing from $57.9$ to $51.4$, while agent-based methods further lag behind, dropping from $47.3$ to $41.2$. 
These results suggest that increasing structural complexity primarily exposes the long-form controllability gap between model families.
Event-chain length provides a complementary measure of long-form pressure. 
As shown in Figure~\ref{fig:event_count_family_bars}, commercial models remain comparatively robust when moving from shorter to longer event chains, with scores decreasing only from $76.0$ to $74.4$. 
In contrast, open-source models drop more substantially from $59.4$ to $52.0$, while agent-based methods remain much lower and decrease from $46.2$ to $44.6$. Overall, both difficulty level and event-chain length reveal that current models degrade as long-form generation requires more events, stronger temporal organization, and more demanding scene structure.

\paragraph{A Shared Failure Mode Across Models.}
Among all application scenarios, \textit{Performance Ads} emerges as the most challenging setting for current models. 
We observe that it is the scenario in which the largest number of systems achieve their lowest overall scores. 
Further analysis shows that the performance degradation is primarily driven by drops in event fulfillment ($\mathrm{V}_{\mathrm{QA}}$) and long-form continuity (Cont.). 
This trend is especially pronounced for open-source models, whose declines in these two dimensions are substantially larger than in other scenarios. 
These results suggest that current systems are not failing to understand product-oriented prompts at the semantic level; rather, they struggle to reliably execute product presentation, functional demonstration, causal progression, and multi-step selling-point delivery over extended durations. 
Case analysis further reveals recurring failure patterns, including missing product operations, broken demonstration sequences, inconsistent causal outcomes, and unstable narrative pacing. 
Overall, the challenges exposed by this scenario are fundamentally associated with physical process modeling and commercial narrative organization, rather than simple text-video semantic alignment.

\subsection{Human Alignment}
\begin{figure}[t]
  \centering
  \includegraphics[width=\columnwidth]{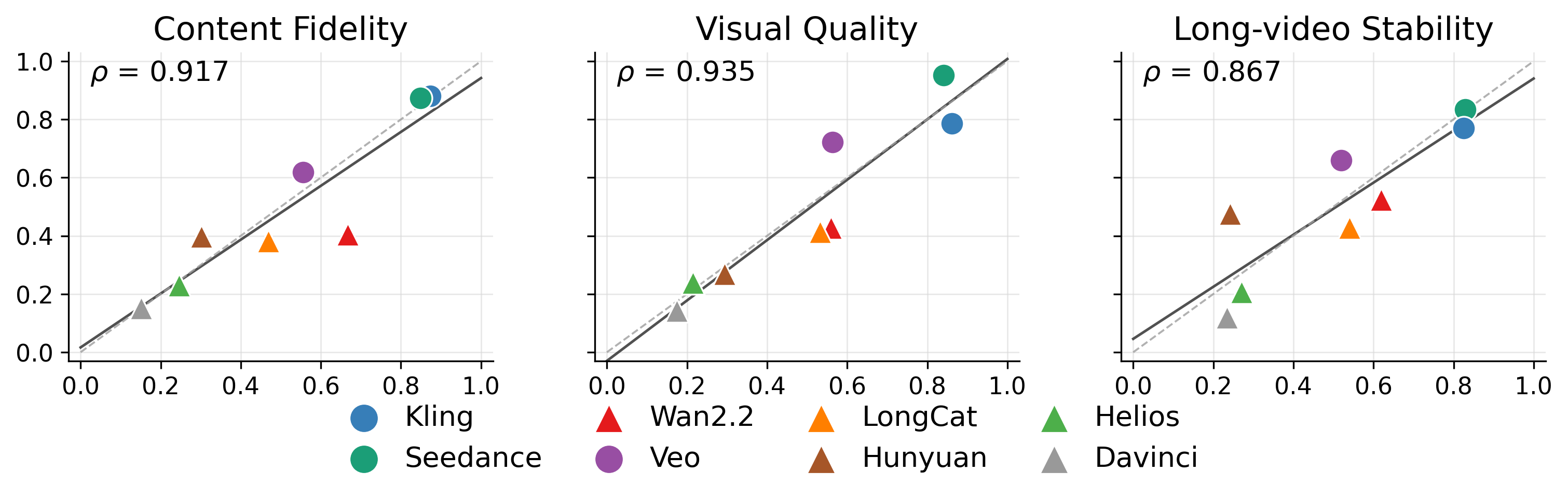}
  \caption{\textbf{Human-alignment validation.} Each point denotes one model,
  with proprietary models shown as circles and open-source models shown
  as triangles. The three panels compare human-derived and
  benchmark-derived pairwise win rates for content fidelity, visual
  quality, and long-video stability.}
  \label{fig:human_alignment_winrate}
\end{figure}

To validate whether the automatic scores reflect human preferences, we conduct a pilot human-alignment study on $40$ selected cases. 
Human raters evaluate each output along three dimensions: content fidelity, visual quality, and long-video stability. 
We align these dimensions with benchmark metrics by aggregating event fulfillment and text-video alignment for content fidelity, event-level visual quality, and holistic presentation for visual quality, and long-form structure and transition stability for long-video stability. 
Following prior preference-based validation protocols, we convert both human ratings and benchmark scores into pairwise outcomes within each sample, where Win=$1$, Loss=$0$, and Tie=$0.5$. 
We then average these outcomes into model-level win rates and compute the Pearson correlation between human-derived and benchmark-derived win rates. 
As shown in Figure~\ref{fig:human_alignment_winrate}, \name achieves strong alignment with human preferences, with Pearson correlations of $0.917$ for content fidelity, $0.935$ for visual quality, and $0.867$ for long-video stability. 
These results suggest that \name captures human preferences over content completion, visual generation quality, and long-form stability, while serving as a pilot validation rather than a replacement for large-scale human evaluation.

\begin{table}[t]
\centering
 \caption{\textbf{Input-format sensitivity analysis.} Each entry reports the average balanced score
  obtained by generating from the same source content under different conditioning formats: V2AV,
  I2AV, and T2AV.}
\label{tab:task_preference_by_model}
\begin{tabular}{lccc}
\toprule
\textbf{Model} & \textbf{V2AV} & \textbf{I2AV} & \textbf{T2AV} \\
\midrule
Seedance 2.0    & 80.4 & 83.9 & \textbf{83.6} \\
Veo 3.1          & 57.4 & \textbf{71.8} & 68.1 \\
LongCat      & 39.8 & 40.4 & \textbf{41.2} \\
Helios (14B)  & \textbf{40.5} & 34.4 & 34.6 \\
\bottomrule
\end{tabular}
\end{table}

\subsection{Input Format Sensitivity in Long-Video Generation}

To examine how input format affects long-video generation, we construct multiple input variants from transcripts of the same real video and generate outputs with different models. 
The results are reported in Table~\ref{tab:task_preference_by_model}. 
Even under the same source content, models exhibit clear differences in their adaptability to different conditioning formats. 
Notably, although V2AV provides the richest reference information, it does not consistently yield the best long-video outputs. 
Instead, the optimal input format is often model-dependent. 
For example, Helios achieves higher-quality long videos under the V2AV setting, whereas Veo produces more stable results under the I2AV setting. 
These findings suggest that long-video generation does not admit a universally optimal conditioning format. 
In practice, the input formulation and generation strategy should be selected according to each model's conditioning interface and generation behavior to improve both output quality and temporal stability.

\subsection{Reproducibility and Release Plan}

Reproducibility is treated as an integral part of the benchmark design. 
For all MLLM-based evaluations, we record the exact model version and API snapshot time to account for possible changes in commercial systems. 
We will release the task annotations, raw MLLM JSON outputs, evaluation scripts, and aggregate score files together with the benchmark. 
This release will allow future work to audit benchmark judgments, recompute results, and reuse evaluation traces without rerunning the full evaluation pipeline from scratch.

\section{Conclusion}
\label{sec:conclusion}

We introduce \name, a unified benchmark for minute-scale audio-visual generation across text, image, and video inputs. 
By combining taxonomy-guided test construction with a task-aligned diagnostic evaluation framework, \name moves evaluation beyond short clips toward structured long-form scenarios that modern generation systems increasingly target. 
Our evaluation of representative systems shows that current models cannot be adequately characterized by a single overall score: strong long-form generation requires event completion, temporal continuity, visual quality, semantic alignment, and audio-visual synchronization to hold jointly over extended durations. 
Further analysis reveals common bottlenecks in product-oriented scenarios, degradation under increasing event complexity, model-dependent sensitivity to input format, and gaps between native audio support and reliable long-form audio generation. 
These findings position \name not only as a benchmark for comparing systems, but also as a diagnostic testbed for identifying where current audio-visual generation models fail as temporal scope, conditioning diversity, and cross-modal coupling become more demanding.

{
    \small
    \bibliographystyle{ieeenat_fullname}
    \bibliography{main}
}

\clearpage
\appendix



\section{Data Construction Details}
\label{app:data_construction}

\subsection{Prompt Design Templates}
\label{app:prompt_templates}

Our benchmark constructs structured long-form scripts through multiple
pipelines, each using Gemini 3.1 Pro with tailored prompts. We detail
the prompt design for each construction track below.

\paragraph{T2AV Real-Video Transcription.}
For the real-video track, we provide Gemini 3.1 Pro with a source video
and instruct it to produce a structured script. The prompt template
(Figure~\ref{fig:prompt_t2av_real}) requires the model to decompose the
video into semantically coherent events, each containing 2--4 shots
with detailed visual descriptions, audio expectations, and a
verifiable completion flag. The output also includes
\texttt{identity\_tracking} for recurring subjects and
\texttt{physical\_constraints} for scene consistency.

\paragraph{T2AV LLM-Template Generation.}
For the LLM-generated track, human designers specify scenario,
complexity level (L1--L4), and language, then Gemini 3.1 Pro generates
the full script following complexity guidelines
(Figure~\ref{fig:prompt_t2av_llm}). Each event includes 3 fixed QA
questions covering subject presence, core action occurrence, and key
visual detail correctness.

\paragraph{I2AV Image-Conditioned Generation.}
For the I2AV LLM-template track, we first extract a structured
\texttt{image\_prior} from the reference image
(Figure~\ref{fig:prompt_i2av}), capturing subjects, objects,
composition, lighting, motion potential, and consistency constraints.
The extracted prior is then fed into a second prompt that generates the
full event script while preserving visual anchoring to the reference
image.

\paragraph{V2AV Continuation Construction.}
For V2AV, each case provides a 10--15\,s reference video. Gemini 3.1
Pro watches the reference clip and generates a continuation script for
the remaining 45--50\,s (Figure~\ref{fig:prompt_v2av}). The output
includes separate \texttt{video\_prompt} and \texttt{audio\_prompt}
fields that concatenate all event descriptions for downstream
generation models.


\begin{figure}[t]
\begin{tcolorbox}[colback=softblue,colframe=cvprblue,title=T2AV Real-Video Transcription Prompt,fonttitle=\small\bfseries,top=3pt,bottom=3pt,left=4pt,right=4pt]
\scriptsize
\texttt{You are a professional video script analyst. Watch the provided
video carefully and produce a structured long-form script in the
following JSON format:}\\[2pt]
\texttt{\{} \\
\texttt{~~"language": "zh|en",} \\
\texttt{~~"global\_description": "<complete narrative summary>",} \\
\texttt{~~"events": [\{} \\
\texttt{~~~~"event\_id": 1,} \\
\texttt{~~~~"time\_range": "0-Xs",} \\
\texttt{~~~~"action": "<one-sentence summary>",} \\
\texttt{~~~~"completion\_flag": "<observable end condition>",} \\
\texttt{~~~~"visual\_description": "<detailed visual content>",} \\
\texttt{~~~~"audio\_expectation": "<expected audio content>",} \\
\texttt{~~~~"shot1": "<shot description>",} \\
\texttt{~~~~"shot1\_time\_range": "0-Ys",} \\
\texttt{~~~~...} \\
\texttt{~~\}]} \\
\texttt{\}}\\[2pt]
\texttt{Requirements:} \\
\texttt{- Decompose the video into semantically coherent events.} \\
\texttt{- Each event should contain 2-4 shots.} \\
\texttt{- Describe visual content in detail: subjects, actions, camera, lighting.} \\
\texttt{- Specify expected audio: dialogue, music, sound effects, ambience.} \\
\texttt{- Include identity\_tracking for recurring subjects.} \\
\texttt{- Include physical\_constraints and narrative\_logic.}
\end{tcolorbox}
\caption{Prompt template for T2AV real-video transcription.}
\label{fig:prompt_t2av_real}
\end{figure}

\begin{figure}[t]
\begin{tcolorbox}[colback=softgreen,colframe=green!50!black,title=T2AV LLM-Template Generation Prompt,fonttitle=\small\bfseries,top=3pt,bottom=3pt,left=4pt,right=4pt]
\scriptsize
\texttt{Generate a structured video script for a \{scenario\} video at
complexity level \{level\} in \{language\}.}\\[2pt]
\texttt{Complexity guidelines:} \\
\texttt{- L1: Multiple entities or simple short-range interactions} \\
\texttt{- L2: Multi-event structures and cross-event transitions} \\
\texttt{- L3: Multi-actor interactions, role consistency, and longer-range dependencies} \\
\texttt{- L4: Causal chains, physical plausibility, and demanding story closure}\\[2pt]
\texttt{Output the same JSON schema as above, including:} \\
\texttt{- global\_description covering the entire 60s video} \\
\texttt{- events with shots, QA questions, and audio expectations} \\
\texttt{- identity\_tracking for all recurring subjects} \\
\texttt{- physical\_constraints and narrative\_logic}\\[2pt]
\texttt{Each event must include 3 fixed QA questions covering:} \\
\texttt{1. subject: Is the main subject/product present?} \\
\texttt{2. action/interaction: Does the core action occur?} \\
\texttt{3. attribute/scene/text\_brand: Are key visual details correct?}
\end{tcolorbox}
\caption{Prompt template for T2AV LLM-template generation.}
\label{fig:prompt_t2av_llm}
\end{figure}

\begin{figure}[t]
\begin{tcolorbox}[colback=softyellow,colframe=orange!60!black,title=I2AV Image Prior Extraction Prompt,fonttitle=\small\bfseries,top=3pt,bottom=3pt,left=4pt,right=4pt]
\scriptsize
\texttt{Analyze this image and produce a structured prior in JSON:}\\[2pt]
\texttt{\{} \\
\texttt{~~"image\_summary": "<one-sentence summary>",} \\
\texttt{~~"detailed\_visual\_description": "<full visual analysis>",} \\
\texttt{~~"subjects": [...],} \\
\texttt{~~"key\_objects": [...],} \\
\texttt{~~"environment\_and\_setting": "<scene description>",} \\
\texttt{~~"composition\_and\_camera": "<framing and angle>",} \\
\texttt{~~"lighting\_and\_color": "<lighting and palette>",} \\
\texttt{~~"visible\_actions\_or\_states": [...],} \\
\texttt{~~"visible\_text\_or\_branding": "<text/logo if any>",} \\
\texttt{~~"inferred\_context": "<likely use case>",} \\
\texttt{~~"motion\_potential": ["<possible video continuations>"],} \\
\texttt{~~"audio\_implications": ["<likely audio content>"],} \\
\texttt{~~"consistency\_constraints": ["<what must stay stable>"],} \\
\texttt{~~"uncertain\_points": ["<ambiguous elements>"]~\}}
\end{tcolorbox}
\caption{Prompt template for I2AV image prior extraction.}
\label{fig:prompt_i2av}
\end{figure}

\begin{figure}[t]
\begin{tcolorbox}[colback=softpurple,colframe=purple!60!black,title=V2AV Continuation Prompt,fonttitle=\small\bfseries,top=3pt,bottom=3pt,left=4pt,right=4pt]
\scriptsize
\texttt{Watch the reference video and generate a natural continuation
script. The continuation should:}\\[2pt]
\texttt{- Maintain subject identity, style, and setting from the reference} \\
\texttt{- Extend the narrative logically for 45-50 additional seconds} \\
\texttt{- Decompose into 4-7 events with visual and audio descriptions} \\
\texttt{- Include identity\_tracking and physical\_constraints}\\[2pt]
\texttt{Additionally, produce:} \\
\texttt{- video\_prompt: all event visual\_descriptions joined by semicolons} \\
\texttt{- audio\_prompt: all event audio\_expectations joined by semicolons} \\
\texttt{- speech\_prompt: any required spoken text or titles}
\end{tcolorbox}
\caption{Prompt template for V2AV continuation construction.}
\label{fig:prompt_v2av}
\end{figure}

\subsection{Dataset Statistics}
\label{app:dataset_statistics}

\begin{figure*}[t]
\centering
\includegraphics[width=\linewidth]{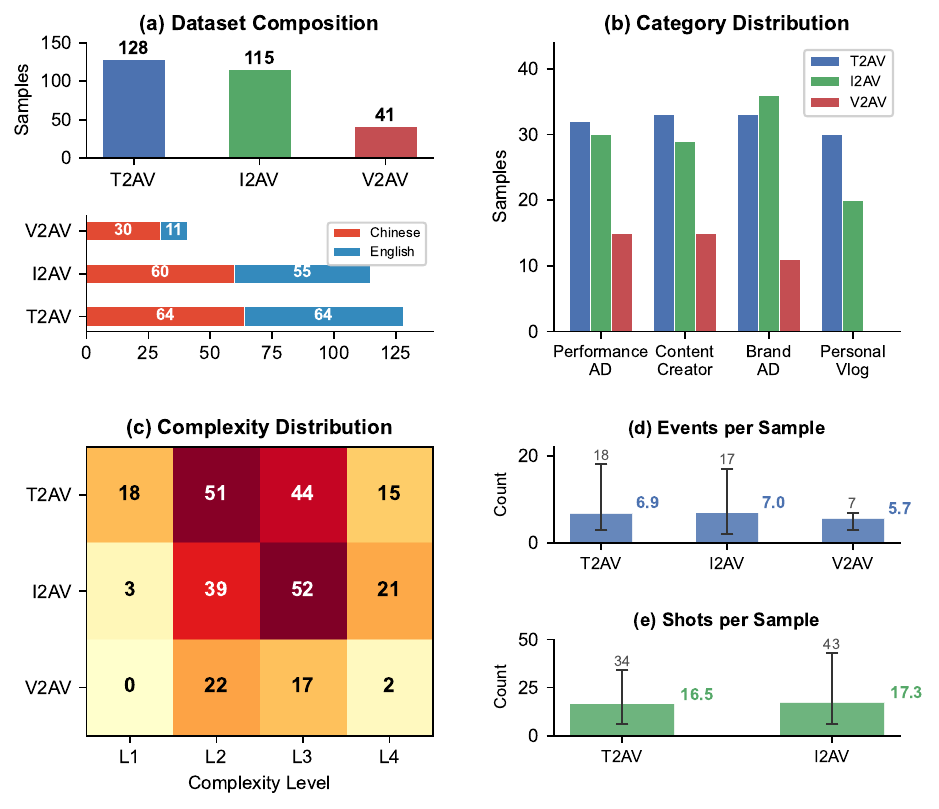}
\caption{\textbf{Dataset statistics of \name.} (a)~Sample count per task with language distribution below. (b)~Category distribution across tasks. (c)~Complexity level distribution (L1--L4) per task. (d)~Events per sample (bar = mean, error bar = min/max range). (e)~Shots per sample.}
\label{fig:dataset_stats}
\end{figure*}

Figure~\ref{fig:dataset_stats} provides a comprehensive overview of the
dataset statistics. LongAV-Compass contains 284 samples across three
tasks, four application scenarios, and
four complexity levels (L1--L4). As shown in panel~(a), T2AV is the
largest split with balanced Chinese/English coverage, while V2AV
focuses on Chinese-dominant continuation scenarios. Panel~(b)
demonstrates a balanced distribution across the four application
scenarios (Performance Ads, Content-Creator, Brand Ads, Vlog).
The complexity heatmap in panel~(c) reveals that L2--L3 dominate across
all tasks, with V2AV containing no L1 samples due to the inherent
complexity of video continuation. Panels~(d) and~(e) show that T2AV
and I2AV have similar event counts (avg 6.9--7.0, range 2--18) and shot
counts (avg 16.5--17.3), while V2AV is more constrained (avg 5.7
events, range 3--7).

\section{Evaluation Framework Details}
\label{app:eval_framework}

\subsection{Evaluation Dimensions and Scoring Rubric}
\label{app:scoring_rubric}

Our evaluation framework assesses generated videos along six shared
video dimensions and three audio dimensions. Each dimension uses
sub-item decomposition. All MOS scores use a 1--5 scale with the
following anchors: 1=failed/severe defects, 2=poor with major issues,
3=acceptable with visible issues, 4=good with minor issues,
5=excellent.

\paragraph{Video Evaluation Dimensions.}
Table~\ref{tab:video_rubric} details the six video evaluation dimensions
and their constituent sub-items.

\begin{table*}[t]
\caption{Video evaluation dimensions with sub-items.}
\label{tab:video_rubric}
\centering
\small
\begin{tabular}{lll}
\toprule
\textbf{Dimension} & \textbf{Sub-item} & \textbf{Definition} \\
\midrule
\multirow{3}{*}{\shortstack[l]{Event Fulfillment\\($V_\mathrm{QA}$)}}
& Fixed QA checklist & Each question: yes=1.0, partial=0.5, no=0.0 \\
& Event score & Average of QA scores within event \\
& Sample score & Duration-weighted average across events \\
\midrule
\multirow{4}{*}{\shortstack[l]{Event Realization\\(VQ)}}
& motion\_naturalness & Motion fluidity, no unnatural jumps \\
& subject\_integrity & Subject structure stability, no deformation \\
& artifact\_control & No ghosting, flickering, broken geometry \\
& visual\_quality & Frame-level clarity and local quality \\
\midrule
\multirow{4}{*}{\shortstack[l]{Long-form Structure\\(Cont.)}}
& event\_order\_correctness & Events follow the prescribed sequence \\
& coverage\_balance & No event is skipped or excessively prolonged \\
& pacing\_consistency & Rhythm is smooth, no sudden stalls \\
& cross\_event\_continuity & Events form a coherent whole, not fragments \\
\midrule
\multirow{2}{*}{\shortstack[l]{Transition Stability\\(Trans)}}
& Algorithm score & Black frames, flashes, duplicates, freezes \\
& LLM score & Generation breaks, deformation, disappearance \\
\midrule
\multirow{4}{*}{\shortstack[l]{Holistic Presentation\\(Hol.)}}
& style\_consistency & Unified visual style throughout \\
& visual\_appeal & Overall aesthetic quality \\
& commercial\_completeness & Feels like a complete production \\
& overall\_watchability & Engaging viewing experience \\
\midrule
\multirow{1}{*}{\shortstack[l]{Text-Video Alignment\\(TVAlign)}}
& Global alignment score & 0.0--1.0 semantic match to prompt \\ \\
\bottomrule
\end{tabular}
\end{table*}

\paragraph{Audio Evaluation Dimensions.}
Table~\ref{tab:audio_rubric} lists the three audio evaluation dimensions.

\begin{table*}[t]
\caption{Audio evaluation dimensions with sub-items.}
\label{tab:audio_rubric}
\centering
\small
\begin{tabular}{lll}
\toprule
\textbf{Dimension} & \textbf{Sub-item} & \textbf{Definition} \\
\midrule
\multirow{2}{*}{\shortstack[l]{Audio-Video Sync\\(AVS)}}
& LLM av\_sync score & Sound aligns with visible actions and cuts \\
& Algorithm sync score & Motion-audio energy peak correlation \\
\midrule
\multirow{3}{*}{\shortstack[l]{Audio Quality\\(AudQ)}}
& audio\_event\_match & Audio matches event text and audio expectation \\
& audio\_realism & Natural, clear, plausible for the scene \\
& audio\_artifact\_control & No clipping, buzzing, glitches, loops \\
\midrule
\multirow{4}{*}{\shortstack[l]{Audio Long-range\\Coherence (AudL)}}
& audio\_continuity & No unexplained dropouts or breaks \\
& ambience\_stability & Background music/ambience stays coherent \\
& source\_consistency & Voices and sound sources stay plausible \\
& volume\_stability & No abrupt volume jumps or distortion \\
\bottomrule
\end{tabular}
\end{table*}

\paragraph{Task-Specific Dimensions.}
\begin{itemize}[nosep]
\item \textbf{I2AV -- First-frame Anchoring ($IV_1$)}: Gemini 1--5 MOS
  rating of whether the generated video opening preserves the reference image.
\item \textbf{I2AV -- Image Alignment (ImgAlign)}: CLIP ViT-L/14
  cosine similarity between the reference image embedding and
  uniformly sampled event frames, aggregated via trimmed mean per
  event and duration-weighted average.
\end{itemize}

\section{Case Studies}
\label{app:case_study}

\begin{figure*}[t]
\centering
\includegraphics[width=\linewidth]{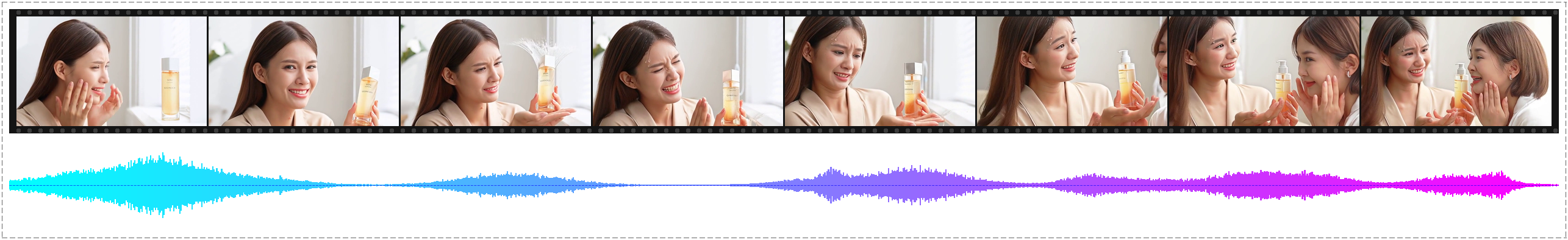}
\caption{Keyframe visualization of the T2AV Performance Ads case (L4). The filmstrip shows representative frames from each event, illustrating the two-actor skincare demonstration sequence.}
\label{fig:case_t2av}
\end{figure*}

\begin{figure*}[t]
\centering
\includegraphics[width=\linewidth]{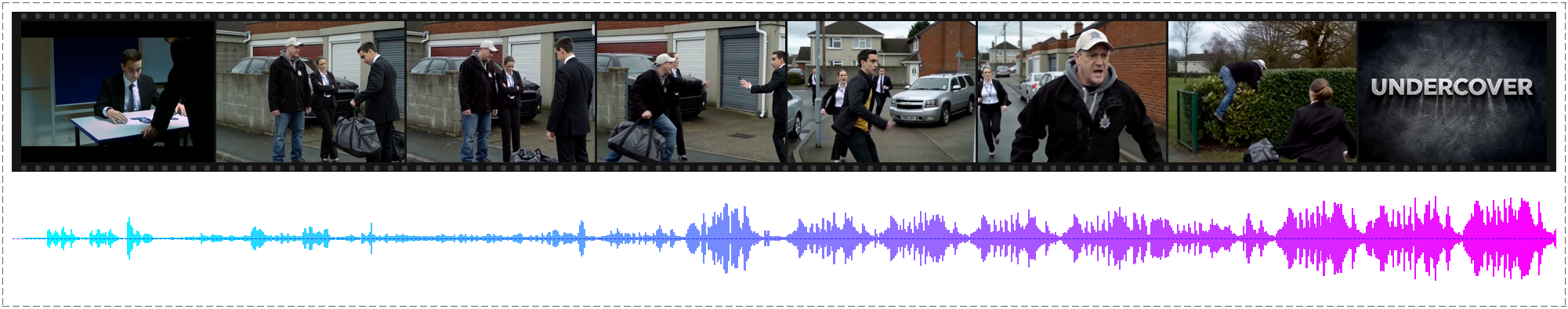}
\caption{Keyframe visualization of the V2AV Content-Creator case (L4). The continuation begins after the reference video, showing the collision, eye contact, romantic montage, and bittersweet ending.}
\label{fig:case_v2av}
\end{figure*}

\begin{figure*}[t]
\centering
\includegraphics[width=\linewidth]{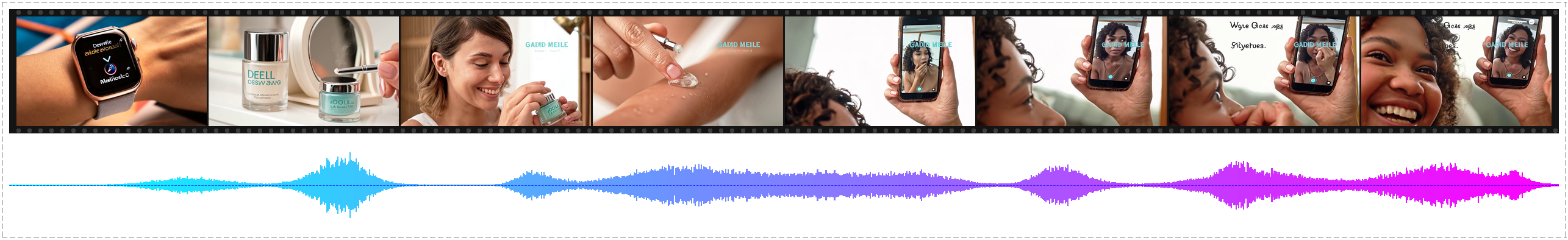}
\caption{Keyframe visualization of the I2AV Performance Ads case (L4). Starting from the reference product image (leftmost frame), the generated video demonstrates the Apple Watch through lifestyle scenarios while preserving product appearance.}
\label{fig:case_i2av}
\end{figure*}

This section presents representative benchmark cases to illustrate the
annotation quality, generation challenges, and evaluation behavior of
LongAV-Compass. Figures~\ref{fig:case_t2av}, \ref{fig:case_v2av}, and
\ref{fig:case_i2av} show keyframe visualizations for the three cases.

\subsection{T2AV Case: Performance Ads Product Demonstration (L4)}
\label{app:case_t2av}

\paragraph{Overview.}
This LLM-generated case targets a 60-second skincare product
advertisement with two actors, four events, and rich
product-interaction sequences. Complexity level L4 requires multi-actor
coordination and role consistency.

\paragraph{Global Description.}
\begin{quote}\small
At the beginning, Xiao Ya (Main Character 2) sits by a bright window, gently stroking her cheek with a troubled expression. The camera focuses on a close-up of her face, hinting at dry skin. At this moment, Xiao Lin (Main Character 1) approaches from off-screen, with a confident smile, holding a bottle of sophisticatedly simple "Water Glow Serum" in front of Xiao Ya. The camera zeroes in on Xiao Lin's hand movements. She elegantly twists off the cap, presses the pump, and squeezes a drop of clear serum onto her palm. A close-up follows as she gently spreads it with her fingertips, and the serum instantly "explodes" into a fine mist of water droplets. Xiao Lin turns to Xiao Ya, indicating with her eyes for her to extend her hand. Xiao Ya hesitates but eventually reaches out. Xiao Lin smiles as she squeezes the essence onto Xiao Ya's palm. When Xiao Ya feels the "water explosion" effect, her eyes widen instantly, revealing a surprised smile. The camera finally focuses on a close-up of the product, with the bottle reflecting a soft glow.
\end{quote}

\paragraph{Event Structure.}
\begin{itemize}[nosep]
\item \textbf{Event 1} (0--18s, 3 shots): Xiao-Lin recommends the product to troubled Xiao-Ya.
\item \textbf{Event 2} (18--35s, 3 shots): Xiao-Lin demonstrates the ``water-burst'' texture on her own hand.
\item \textbf{Event 3} (35--50s, 3 shots): Xiao-Ya tries it herself and reacts with surprise.
\item \textbf{Event 4} (50--60s, 2 shots): Product hero shot with brand reveal.
\end{itemize}

\paragraph{Identity Tracking.}
\begin{itemize}[nosep]
\item \texttt{subject\_1}: Xiao-Lin, confident skincare expert in light casual clothing.
\item \texttt{subject\_2}: Xiao-Ya, gentle personality, casual home wear.
\end{itemize}

\paragraph{QA Checklist (Event 1 Example).}
\begin{enumerate}[nosep]
\item (\texttt{subject}) Are two women and a skincare product visible?
\item (\texttt{interaction}) Does one woman hand the product to the other?
\item (\texttt{scene}) Is the setting a bright indoor window area?
\end{enumerate}

\paragraph{Generation Challenges.}
This case tests: (1) two-person interaction with distinct roles,
(2) product close-up with texture demonstration (``water-burst''
effect), (3) facial expression transitions, and (4) brand packaging
consistency across shots.

\subsection{V2AV Case: Content-Creator Short Film Continuation (L4)}
\label{app:case_v2av}

\paragraph{Overview.}
This case requires continuing a short-film narrative from a reference
video. The reference shows a man running through a train station; the
continuation covers a dramatic encounter, fantasy montage, and
bittersweet ending across 6 events.

\paragraph{Global Description (English).}
\begin{quote}\small
After the reference video, the video continues with: The man who is
running collides with another individual, making his briefcase spring
open. White sheets of paper fly out, scattering over the tiled
platform. Kneeling to gather his papers, the man glances up and
establishes direct eye contact with a dark-haired woman in a black
leather jacket and red lipstick. A quick, dream-like montage depicts
the man and woman in several joyful, romantic situations: hugging,
proposal by a waterfall, mountaintop wedding, holding keys to a new
house, and hand on her pregnant stomach. The footage cuts back to the
platform. The woman boards the train. The man stands alone as the train
departs. Title card: ``THE MISSED CONNECTION.''
\end{quote}

\paragraph{Event Structure.}
\begin{itemize}[nosep]
\item \textbf{Event 2} (8--13s): Collision and paper scatter.
\item \textbf{Event 3} (13--17s): Eye contact between man and woman.
\item \textbf{Event 4} (17--32s): Dream-like romantic montage (6 sub-scenes).
\item \textbf{Event 5} (32--40s): Return to reality; woman boards train.
\item \textbf{Event 6} (40--52s): Man rises; train departs; alone on platform.
\item \textbf{Event 7} (52--65s): Title card and credits.
\end{itemize}

\paragraph{Identity Tracking.}
\begin{itemize}[nosep]
\item \texttt{subject\_1}: Young Caucasian man with reddish-blond hair,
  dark fitted business suit, white shirt, brown shoes, brown leather
  briefcase.
\item \texttt{subject\_2}: Young Caucasian woman with long dark brown
  hair, black leather jacket, white cargo pants, black boots, red
  lipstick, checkered tote bag.
\end{itemize}

\paragraph{Physical Constraints.}
\begin{itemize}[nosep]
\item Both subjects must maintain facial/physical features throughout all scenes including the montage.
\item The train platform is an elevated outdoor station with modern architecture.
\item The man does not board the train.
\item Scattered papers and open briefcase remain on the platform at the end.
\end{itemize}

\paragraph{Generation Challenges.}
This case is particularly demanding because it requires: (1)
continuation consistency with the reference video (style, subject),
(2) a fantasy montage with multiple rapid scene changes while
preserving identity, (3) emotional transitions (surprise $\to$ love
$\to$ loss), and (4) cross-event props consistency (papers on floor).

\subsection{I2AV Case: Product Lifestyle Image to Video (L4)}
\label{app:case_i2av}

\paragraph{Overview.}
Starting from a product flat-lay photo of an Apple Watch on a wooden
desk, the model must generate a 60-second performance advertisement video that
preserves the watch's appearance and desktop environment.

\paragraph{Image Prior (extracted fields).}
\begin{itemize}[nosep]
\item \textbf{Image Summary}: A product lifestyle shot of an Apple Watch
  with a woven band, alongside a notebook, pen, and iPod on a wooden desk.
\item \textbf{Key Objects}: Apple Watch (silver aluminum), woven sport
  band (gray-white), blue notebook, white stylus, white iPod Classic.
\item \textbf{Composition}: Close-up from above at an angle, shallow
  depth of field, watch as visual center.
\item \textbf{Lighting}: Warm directional light from left, highlights on
  metal frame.
\item \textbf{Consistency Constraints}: Watch must remain silver
  aluminum; band must stay gray-white woven; desk environment must
  persist; warm soft lighting maintained.
\end{itemize}

\paragraph{Generated Script (5 events).}
\begin{enumerate}[nosep]
\item (0--9s) Camera establishes the desk scene; watch receives a notification.
\item (9--20s) A hand enters frame, picks up the watch, puts it on wrist.
\item (20--35s) Close-up: watch face shows fitness rings filling; demonstrates health tracking.
\item (35--48s) Quick-cut montage of different watch faces and complications.
\item (48--60s) Return to desk scene; brand logo and CTA appear.
\end{enumerate}

\paragraph{Generation Challenges.}
This case tests: (1) precise preservation of the reference image's
visual style and objects, (2) natural transition from still product
shot to dynamic video, (3) fine-grained UI animation on the watch
screen, and (4) return to the original composition for the ending.

\subsection{Challenging Cases}
\label{app:hard_cases}

We identify several categories of cases that are particularly
difficult for current generation models:

\paragraph{Case 1: High Event Count (18 events).}
A real-video transcription of a product review video
(\texttt{T2AV/Real/performance\_ad}) with 18 events covering nail art
demonstration across multiple colors and techniques. The extreme event
count means models must maintain consistent hand appearance, product
colors, and background across many rapid transitions while following a
complex sequential procedure.

\paragraph{Case 2: Multi-Actor Drama with Emotional Arcs (L4, 13
events).}
A Content-Creator drama (\texttt{T2AV/Real/content\_creator}) where
multiple characters interact with anger, grief, and confrontation. Each
character has distinct costume and facial features that must persist
across 13 events with diverse camera angles. Models typically fail on
identity preservation after event 6--7 or collapse emotional
expression into a neutral default.

\paragraph{Case 3: Aerial Cinematography with Continuous Motion (L4, 15
events).}
A brand advertisement (\texttt{T2AV/Real/brand\_ad}) featuring drone
footage over a desert landscape with continuous camera movement,
text overlays, and transitions between aerial and ground perspectives.
Models struggle with maintaining consistent landscape geometry across
long continuous shots and seamlessly integrating text elements.

\paragraph{Common Failure Patterns.}
Based on evaluation across these challenging cases, we observe:
\begin{itemize}[nosep]
\item \textbf{Identity drift}: Characters gradually change appearance
  after 30--40 seconds, especially hair and clothing details.
\item \textbf{Event collapse}: Later events are skipped or merged when
  the total event count exceeds the model's effective planning horizon.
\item \textbf{Transition artifacts}: Black frames or style jumps appear
  at event boundaries, particularly when adjacent events have very
  different visual content.
\item \textbf{Product inconsistency}: Branded products change shape,
  color, or labeling between shots.
\item \textbf{Audio-visual desynchronization}: For audio-capable models,
  sound effects drift from their visual triggers in later events.
\end{itemize}

\section{Generation Protocol Details}
\label{app:generation_protocol}

\subsection{Prompt Construction for Generation}

When feeding benchmark cases to generation models, we construct
task-specific prompts from the structured annotation.

\paragraph{T2AV Prompt.}
The generation prompt is the \texttt{global\_description} field. For
models that support structured event inputs, we additionally provide
the event list. For models requiring separate audio prompts, we
construct an \texttt{audio\_prompt} by concatenating all event
\texttt{audio\_expectation} fields with semicolons:

\begin{tcolorbox}[colback=lightgray,colframe=gray,fonttitle=\small\bfseries,title=T2AV Generation Input]
\small
\textbf{Video prompt}: \texttt{\{global\_description\}}\\[4pt]
\textbf{Audio prompt} (if supported): \\
\texttt{\{event1.audio\_expectation\}; \{event2.audio\_expectation\}; ...}
\end{tcolorbox}

\paragraph{I2AV Prompt.}
The model receives both a reference image and the
\texttt{global\_description}. For systems that accept structured
prompts, we include the full event list with time ranges:

\begin{tcolorbox}[colback=lightgray,colframe=gray,fonttitle=\small\bfseries,title=I2AV Generation Input]
\small
\textbf{Reference image}: \texttt{\{source\_image\}}\\[4pt]
\textbf{Video prompt}: \texttt{\{global\_description\}}\\[4pt]
\textbf{Audio prompt} (if supported): \\
\texttt{\{event1.audio\_expectation\}; \{event2.audio\_expectation\}; ...}
\end{tcolorbox}

\paragraph{V2AV Prompt.}
The model receives a reference video clip (10--15\,s) plus a
continuation instruction. The \texttt{video\_prompt} and
\texttt{audio\_prompt} fields are pre-constructed in the annotation:

\begin{tcolorbox}[colback=lightgray,colframe=gray,fonttitle=\small\bfseries,title=V2AV Generation Input]
\small
\textbf{Reference video}: \texttt{\{reference\_video.mp4\}} (10--15s)\\[4pt]
\textbf{Video prompt}: ``After the reference video, the scene continues
with: \texttt{\{event2.visual\_description\}}; \texttt{\{event3.visual\_description\}}; ...''\\[4pt]
\textbf{Audio prompt}: ``After the reference video, the audio continues
with: \texttt{\{event2.audio\_expectation\}}; \texttt{\{event3.audio\_expectation\}}; ...''\\[4pt]
\textbf{Speech prompt} (if applicable): \texttt{\{speech\_prompt\}}
\end{tcolorbox}

\subsection{Model-Specific Adaptations}

While preserving event order, conditioning semantics, and audio
expectations, we adapt prompts to each model's native format:

\begin{itemize}[nosep]
\item \textbf{End-to-end models}:
  Receive the full text prompt as a single input; reference
  image/video provided through the model's native conditioning
  interface.
\item \textbf{Pipeline models}:
  Video generation uses the video prompt; audio generation module
  receives the generated video plus the audio prompt.
\item \textbf{Agent-based models}: Receive the event list as structured instructions; the
  agent orchestrates multi-step generation internally.
\item \textbf{Open-source models}: Receive the global description; for models supporting
  only short clips, we use their native multi-segment or autoregressive
  mode to reach the target 60-second duration.
\end{itemize}

\subsection{Output Processing}

Generated videos are post-processed to standardize evaluation:
\begin{enumerate}[nosep]
\item Videos are saved as \texttt{full\_video.mp4} under each model's
  output directory.
\item Event-aligned clips are extracted based on canonical event
  boundaries from \texttt{canonical\_events.json}.
\item Boundary clips (2\,s before and after each event boundary) are
  extracted for transition evaluation.
\item For audio evaluation, audio-bearing event clips are re-extracted
  from \texttt{full\_video.mp4} to ensure audio continuity.
\end{enumerate}

\end{document}